\documentclass{article}

\usepackage{graphicx}
\usepackage{subcaption}
\usepackage{booktabs} 

\usepackage[accepted]{icml2026}

\usepackage{amsmath}
\usepackage{amssymb}
\usepackage{mathtools}
\usepackage{amsthm}
\usepackage{longtable}
\usepackage{booktabs} 
\usepackage{multirow}
\usepackage{makecell}     
\usepackage{array}        
\usepackage{xcolor}      
\usepackage{csquotes}
\usepackage{algorithm}
\usepackage{algorithmic}
\usepackage{amsmath,amssymb}
\usepackage{bbm}
\usepackage{pifont}
\usepackage[table]{xcolor}
\usepackage{xspace}
\newcommand{\eg}{e.g.\@\xspace}

\usepackage{microtype}
\usepackage{hyperref}
\usepackage[normalem]{ulem}
\usepackage[capitalize,noabbrev]{cleveref}

\theoremstyle{plain}

\theoremstyle{definition}

\theoremstyle{remark}

\usepackage{marvosym}

\DeclareRobustCommand{\emailsymbol}{%
  \textsuperscript{\raisebox{-0.15ex}{\scriptsize\Letter}}%
}
\usepackage[textsize=tiny]{todonotes}

\icmltitlerunning{Symbiosis-Inspired Knowledge Distillation for Incremental Object Detection}

\begin{document}

\twocolumn[
  \icmltitle{Symbiosis-Inspired Knowledge Distillation for Incremental Object Detection}



  \icmlsetsymbol{equal}{*}

  \begin{icmlauthorlist}
    \icmlauthor{Mingyue Zeng}{1}
    \icmlauthor{De Cheng\emailsymbol}{1}
    \icmlauthor{Zhipeng Xu}{1}
    \icmlauthor{Huaijie Wang}{2}
    \icmlauthor{Nannan Wang}{1}
    \icmlauthor{Xinbo Gao}{2}
  \end{icmlauthorlist}

  \icmlaffiliation{1}{State Key Laboratory of Integrated Services Networks, School of Telecommunications Engineering, Xidian University, Xi'an, China}
  \icmlaffiliation{2}{School of Electronic Engineering, Xidian University, Xi'an, China}

  \icmlcorrespondingauthor{De Cheng}{dcheng@xidian.edu.cn}

  \icmlkeywords{Machine Learning, ICML}

  \vskip 0.3in
]



\printAffiliationsAndNotice{}  

\begin{abstract}
    Incremental object detection (IOD) aims to extend detectors to new categories while retaining previously acquired knowledge. Existing methods often adopt a class incremental learning perspective, separating feature spaces to sharpen decision boundaries. However, this separation-oriented paradigm may overlook object symbiosis in detection, where co-occurrence and occlusion introduce spatial and semantic dependencies that benefit from shared representations. Ignoring these dependencies distorts the shared representations, exacerbates confusion between old and new classes, and accelerates catastrophic forgetting. To address this, we propose \uline{S}ymbiosis-\uline{I}nspired \uline{K}nowledge \uline{D}istillation (SIKD), which explicitly leverages object symbiosis at two complementary levels. Spatial Symbiosis Distillation (SpSD) focuses on symbiotic regions where the old model responds with high overlap to objects in the new task. It preserves generalizable old class cues, suppresses class-specific bias and redundancy, and distills the refined evidence to the new model at matched spatial locations with slot-aligned supervision. Semantic Symbiosis Distillation (SeSD) maintains class level structure by forming confidence weighted prototypes for old classes and aligning their inter class soft ranks over the old class logits, which stabilizes the semantic topology during adaptation. Extensive experiments demonstrate the effectiveness and superiority of the proposed method.
\end{abstract}
\section{Introduction}
\label{sec:intro}

\begin{figure}[t]
  \centering
   \includegraphics[width=0.9\linewidth]{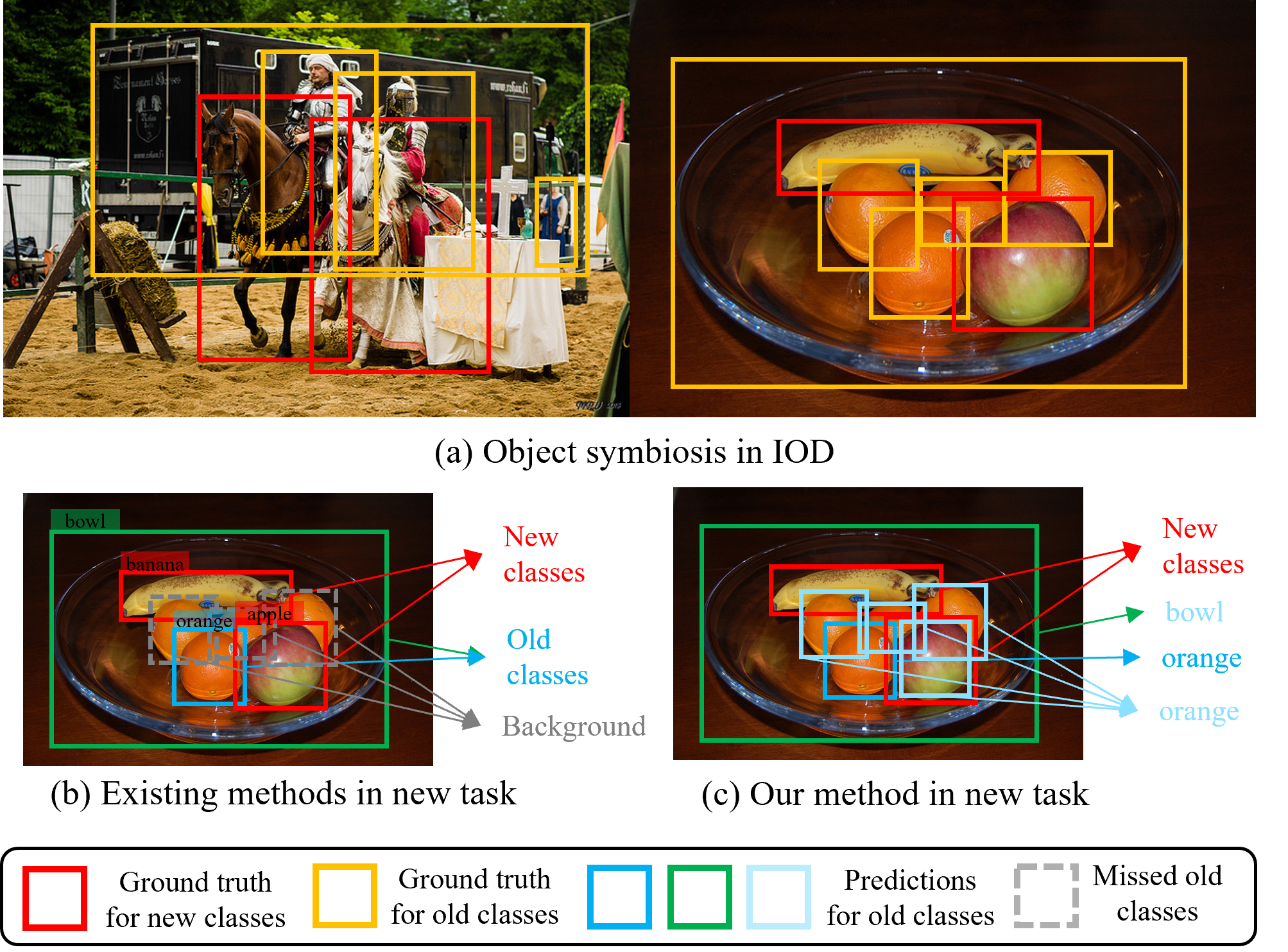}
   \caption{Illustration of (a) object symbiosis in IOD, (b) existing methods in new task, and (c) our method in new task. In (b) and (c), arrows indicate how regional features are classified. In (c), the new-task class \textit{apple} shares coarse features with the old class \textit{orange} while also retaining class-specific cues.}
   \label{fig:intro_figure}
\end{figure}
Object detection has advanced from two-stage frameworks~\cite{girshick2015fast,ren2016faster} to efficient one-stage detectors~\cite{ge2021yolox,tian2019fcos} and end-to-end transformer architectures~\cite{zhu2020deformable,liu2024grounding}. Large-scale pretraining~\cite{dai2021up} and stronger benchmarks have further improved accuracy. However, in real deployments~\cite{cheng2024disentangled,xu2025adversarial}, label spaces evolve as new categories appear. Retraining from scratch for every update is costly and often infeasible when prior data cannot be stored or accessed due to privacy or licensing. Incremental object detection (IOD) addresses this setting by learning new categories while preserving knowledge of learned ones using only the annotations available at each task.

Unlike class incremental classification~\cite{zhou2024class,masana2022class}, where each task provides complete labels for its classes, IOD trains on images that may still contain old objects while only the current categories are annotated. This mismatch pushes unlabeled old-class objects to be treated as background or drift toward new-class labels during training, which accelerates catastrophic forgetting.

To address this issue, existing IOD methods~\cite{liu2023continual,kang2023alleviating,mo2024bridge,kim2024sddgr,wangpsedet,zhang2025dca,wang2025gcd} rely on the old model to mine old-class signals in the current data. As illustrated in Fig.~\ref{fig:intro_figure}(b), they retain only high-confidence old-class detections with low Intersection over Union (IoU) to new-class ground truth. These ``clean'' predictions are treated as the only source of old-task knowledge. This design follows a classification mindset that separates old and new features to reduce entanglement and sharpen decision boundaries~\cite{rebuffi2017icarl,li2024fcs}. However, it overlooks the essential property of object symbiosis in detection. As shown in Fig.~\ref{fig:intro_figure}(a), objects naturally co-occur in shared contexts (\eg, \textit{orange} and \textit{apple} as fruits) and occlude one another (\eg, \textit{person} riding \textit{horse}), which create spatial and semantic dependencies that call for a unified feature space. Filtering supervision to only ``clean'' cases erodes symbiosis-bearing signals and biases the model toward the newly annotated categories, which increases old–new confusion and forgetting.

Instead, we propose Symbiosis-Inspired Knowledge Distillation (SIKD), a framework that maintains a unified feature space by leveraging object symbiosis across both spatial and semantic dimensions. As shown in Fig.~\ref{fig:intro_figure}(c), the old model processing new-task images reveals two symbiotic patterns. Unseen objects are mapped to semantically similar old classes, while partially visible old objects retain detection despite occlusion. We treat these as symbiotic regions encoding shared knowledge rather than noise. By preserving the consistent feature patterns presented in these regions, the model sustains a unified feature space across incremental tasks.

Concretely, SIKD distills symbiotic cues at two levels: instance-level spatial structure and class-level semantic topology, through Spatial Symbiosis Distillation (SpSD) and Semantic Symbiosis Distillation (SeSD). SpSD focuses on co-occurrence and occlusion regions, where it applies a Consistent Feature Enhancement (CFE) module to stabilize overlap-heavy features by reinforcing transferable patterns and suppressing spurious old-class activations. The enhanced features are then distilled to the new model via slot-aligned supervision to preserve spatial dependencies. In parallel, SeSD constructs confidence-weighted prototypes from both symbiotic and non-symbiotic regions and preserves their relative ordering in the old-class subspace via soft rank alignment, thereby maintaining the old-class semantic structure during incremental updates. Together, these two components improve knowledge retention across incremental steps. Our contributions are summarized as follows:
\begin{itemize}
  \item We reinterpret IOD through object symbiosis (co-occurrence and occlusion), exposing the limits of classification-style feature separation in detection.
  \item We propose Symbiosis-Inspired Knowledge Distillation (SIKD), with Spatial Symbiosis Distillation (SpSD) to preserve spatial dependencies in symbiotic regions and Semantic Symbiosis Distillation (SeSD) to preserve old-class semantic topology.
  \item Extensive experiments achieve state-of-the-art performance, and ablations and visual analyses support our method.
\end{itemize}

\begin{figure*}[t]
  \centering
   \includegraphics[width=0.9\linewidth]{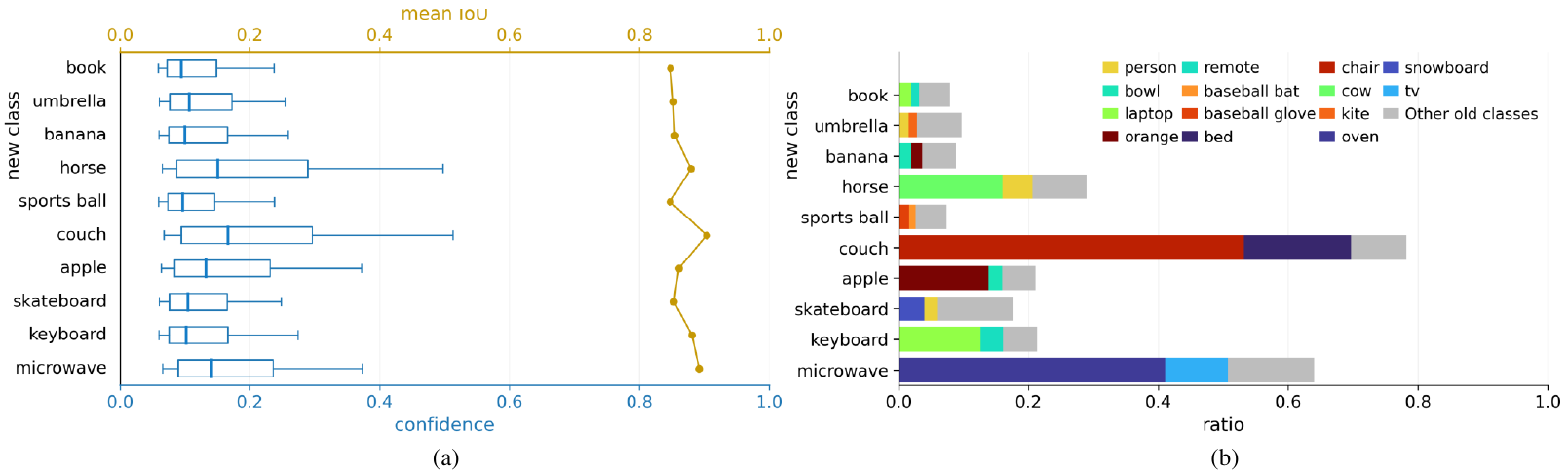}
   \caption{Statistical analysis on COCO 2017 under the 70+10 setting, using old-detector predictions with $\mathrm{IoU}>0.7$ to new-class ground truth. (a) Confidence distribution and mean IoU of the old detector’s old-class predictions on new-class ground-truth instances. (b) Per-class proportion of new-class ground-truth instances misclassified as old classes by the old detector.}
   \label{fig:statistic}
\end{figure*}
\section{Related Work}
\label{sec:Related_work}

\subsection{Incremental Learning}
\label{subsec:Incremental_learning}
Incremental learning aims to acquire new categories over time while preserving prior knowledge, with catastrophic forgetting as the core challenge. Existing methods can be categorized into four main groups. First, output-level distillation~\cite{he2025harnessing,wang2025stpr,rebuffi2017icarl} transfers the old model’s logits and features to the new one to curb prediction drift. Second, parameter regularization~\cite{wang2025ekpc,jung2020continual}, exemplified by elastic weight consolidation~\cite{kirkpatrick2017overcoming}, penalizes changes to important weights so new learning does not overwrite old knowledge. Third, replay or exemplar memory~\cite{aljundi2019online,aljundi2019gradient,zhou2024balanced} stores a small set of representative samples or uses generative replay to stabilize the decision boundary. Fourth, parameter isolation and structural expansion~\cite{he2026task,he2025ckaa,yan2021dynamically,li2019learn} allocate task-specific capacity through masks, sub-networks, or expandable branches to reduce interference between old and new knowledge.

\subsection{Incremental Object Detection}
\label{subsec:Incremental_object_detection}
Incremental Object Detection (IOD) adapts detectors to new categories while retaining previously learned knowledge. Unlike continual learning for classification, where each task uses a fixed label set, IOD operates on images that contain both old and new objects while only the new categories are annotated. This annotation mismatch causes unlabeled old-class instances to be suppressed as background or misassigned to new classes.

Most incremental object detection methods follow a consistent paradigm across different detector architectures. Single-stage detectors like GFL~\cite{li2017learning,li2019rilod,peng2021sid,feng2022overcoming,wangpsedet,zhang2024learning}, two-stage frameworks such as Faster R-CNN~\cite{liu2023augmented,mo2024bridge}, and transformer-based methods~\cite{liu2023continual,kang2023alleviating,zhang2024learning}, all employ pseudo-labeling from previous models to identify old-class instances while filtering out regions potentially containing new categories. In transformer-based methods, CL-DETR~\cite{liu2023continual} selects reliable pseudo labels through dual filtering on IoU and confidence, and BPF~\cite{mo2024bridge} adopts a similar strategy with multiple teachers on Faster R-CNN~\cite{ren2016faster}. Subsequent work enhances this foundation through synthetic exemplar generation using Stable Diffusion \cite{kim2024sddgr} and improved pseudo-label filtering techniques \cite{wangpsedet}. DCA~\cite{zhang2025dca} introduces a localization-then-recognition paradigm that decouples localization from recognition to reduce forgetting, and GCD~\cite{wang2025gcd} incorporates language priors through textual grounding. However, these methods share a fundamental limitation: they treat high-confidence, low-IoU detections as exclusively clean old-class evidence, thereby overlooking the inherent object symbiosis in detection scenarios. In contrast, our SIKD framework explicitly embraces object symbiosis, maintaining a unified feature space and modeling inter-object dependencies rather than suppressing them.
\section{Methodology}
\label{sec:Methodology}
\begin{figure*}[t]
  \centering
  \includegraphics[width=0.95\linewidth]{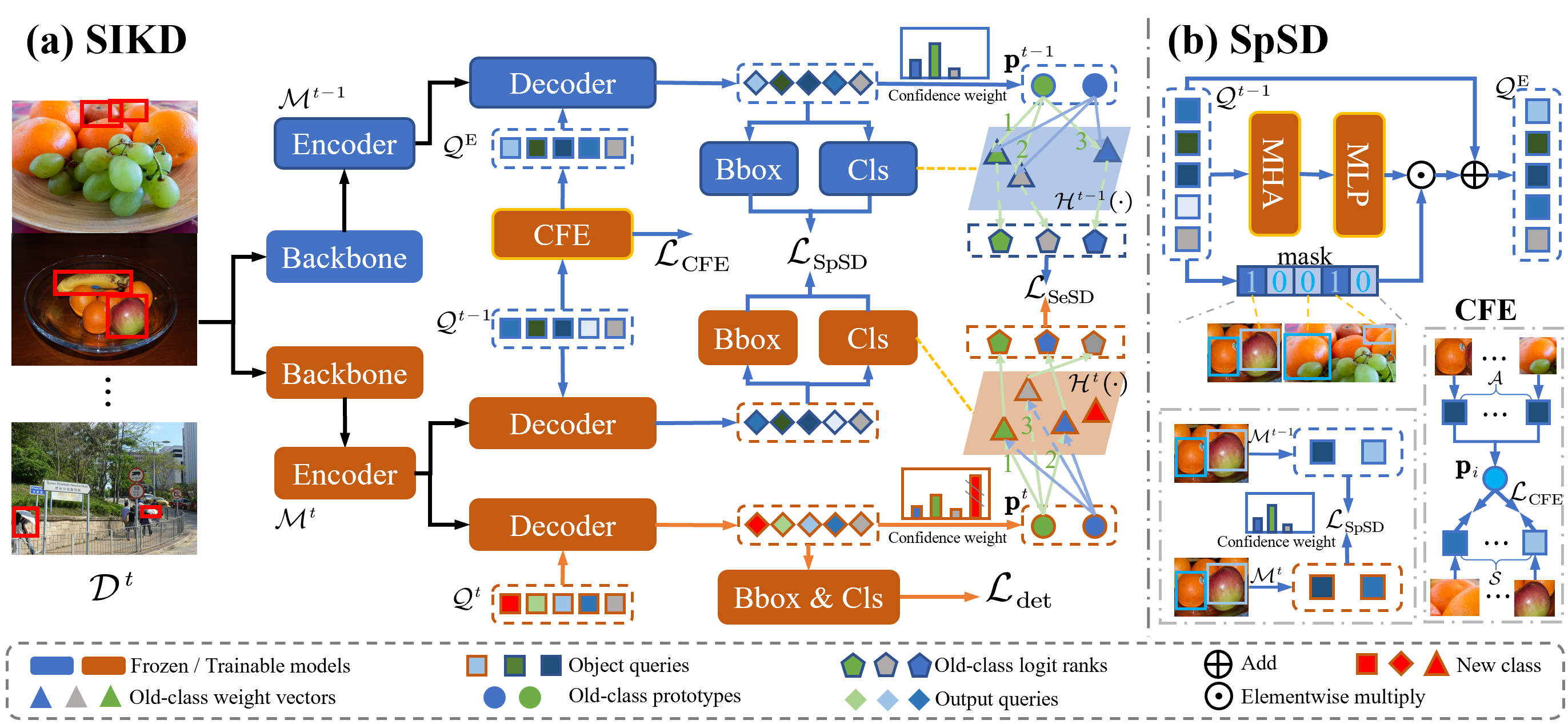}
  \caption{Overview of our proposed SIKD. (a) Training pipeline. The frozen old model $\mathcal{M}^{t-1}$ produces queries $\mathcal{Q}^{t-1}$ on $\mathcal{D}^t$. CFE refines queries of symbiotic regions under anchor-prototype guidance, yielding $\mathcal{Q}^{\mathrm{E}}$ that reduces old-class bias and removes redundancy. SpSD distills anchor logits and boxes, which enforces confidence-weighted, layer-wise logit consistency over all queries. SeSD builds confidence-weighted, $L_2$-normalized old-class prototypes from the last decoder layer of both models and aligns their classifier ranks to preserve the topology of old classes. (b) SpSD module. CFE contains multi-head self-attention (MHA) and an MLP, which is optimized with a prototype-guided cosine loss $\mathcal{L}_\mathrm{CFE}$ on $\mathcal{Q}^{\mathrm{E}}$, and is discarded at inference. 
  }
  \label{fig:overfame}
\end{figure*}
\subsection{Problem Formulation}
\label{subsec:Problem_formulation}
In incremental object detection, the detector is trained over $T$ tasks. The class domain is $\mathcal{C}=\bigcup_{i=1}^{T}\mathcal{C}^i$ with $\mathcal{C}^i\cap\mathcal{C}^j=\emptyset$ for different tasks $i$ and $j$. 
The dataset is $\mathcal{D}=\bigcup_{i=1}^{T}\mathcal{D}^i$, where each $\mathcal{D}^i$ provides annotations $\mathcal{Y}^i$ only for classes in $\mathcal{C}^i$. At task $t$, the model $\mathcal{M}^{t-1}$ is updated to $\mathcal{M}^{t}$ using only $\mathcal{D}^{t}$ and $\mathcal{Y}^{t}$. Images in $\mathcal{D}^{t}$ may still contain unlabeled instances from previously learned classes $\mathcal{C}^{1:t-1}=\bigcup_{i=1}^{t-1}\mathcal{C}^i$. The objective is to learn the new classes $\mathcal{C}^{t}$ while maintaining performance on $\mathcal{C}^{1:t}$ without accessing earlier data $\{\mathcal{D}^{1},\dots,\mathcal{D}^{t-1}\}$.

\subsection{Transformer-based Detectors}
\label{subsec:tbdet}
Following CL-DETR~\cite{liu2023continual}, we adopt Deformable DETR~\cite{zhu2020deformable} as the architecture. In Deformable DETR, a transformer encoder processes image features, and the decoder operates on a set of $n$ learnable object queries $\mathcal{Q}=[\mathbf{q}_1,\dots,\mathbf{q}_n]^\top\!\in\mathbb{R}^{n\times d}$. Each query hypothesizes one object and gathers evidence from the encoded features via cross-attention. A prediction head maps the decoded queries to class logits $\mathbf{z}_i\in\mathbb{R}^e$ and a class-agnostic box $\mathbf{b}_i\in\mathbb{R}^4$, where $e$ is the number of categories.

The decoder of DETR has $L$ layers. At each layer, queries are refined by self-attention, multi-scale deformable cross-attention, and a feed-forward block. Learned reference points are updated across layers, enabling progressive localization and classification refinement. Intermediate predictions are produced at every layer, and the final outputs after $L$ layers are the refined query embeddings $\mathcal{Q}^{(L)}$ together with logits $\{\mathbf{z}_i\}_{i=1}^n$ and boxes $\{\mathbf{b}_i\}_{i=1}^n$.

\subsection{Symbiosis-aware Query Partitioning}
\label{subsec:partition}
As shown in Fig.~\ref{fig:statistic}, incremental object detection naturally exhibits object symbiosis, where old and new categories co-occur and occlude each other in the current training data. When the old model $\mathcal{M}^{t-1}$ processes current data $\mathcal{D}^t$, these relationships emerge as structured patterns in the query space. New objects often activate queries of semantically similar old classes, while partially visible old objects still trigger relevant query responses based on visible cues and context.

At step $t$, we use $\mathcal{M}^{t-1}$ predictions on $\mathcal{D}^t$ to guide the new model $\mathcal{M}^{t}$. Let the old model's queries be $\mathcal{Q}^{t-1}=[\mathbf{q}^{t-1}_1,\dots,\mathbf{q}^{t-1}_n]^\top$ with outputs $\mathbf{z}^{t-1}_i$ for logits and $\mathbf{b}^{t-1}_i$ for the box of each query $\mathbf{q}^{t-1}_i$. The ground truth for $\mathcal{D}^t$ is
\begin{equation}
\mathcal{Y}^t=\{(\mathbf{g}^{t}_j,o^{t}_j)\}_{j=1}^{|\mathcal{Y}^t|},
\label{eq:ground_truth}
\end{equation}
where $\mathbf{g}^{t}_j$ denotes a bounding box with label $o^{t}_j$. 

Let $\sigma(\cdot)$ denote the sigmoid. For each query we define old class confidence and the maximum overlap as
\begin{equation}
\begin{aligned}
s^{t-1}_i &= \max \sigma\!\big(\mathbf{z}^{t-1}_{i}\big),\\
v^{t-1}_i &= \max_{\forall{(\mathbf{g}^{t}_j,o^{t}_j)}\in \mathcal{Y}^t} \mathrm{IoU}\!\big(\mathbf{b}^{t-1}_i,\mathbf{g}^{t}_j\big).
\end{aligned}
\label{eq:conf_and_iou}
\end{equation}

With confidence threshold $\gamma$ and IoU threshold $\tau$, we partition the queries into index sets $\mathcal{A},\mathcal{S},\mathcal{R}\subseteq\{1,\dots,n\}$:
\begin{equation}
\begin{aligned}
\mathcal{A}&=\{\,i \mid s^{t-1}_i\ge\gamma \ \land\  v^{t-1}_i<\tau\,\},\\
\mathcal{S}&=\{\,i \mid v^{t-1}_i\ge\tau\,\},\\
\mathcal{R}&=\{\,i \mid s^{t-1}_i<\gamma \ \land\  v^{t-1}_i<\tau\,\}.
\end{aligned}
\label{eq:ASR_partition}
\end{equation}
This partition makes object symbiosis explicit at the query level where DETR performs inference. The set $\mathcal{A}$ contains high-confidence, low-overlap detections of old classes, which serve as stable anchors. The set $\mathcal{S}$ collects queries from overlapping scenarios caused by co-occurrence and occlusion. It includes cases where unseen new objects are misclassified as semantically similar old classes and cases where partially visible old objects remain detectable. These queries form symbiotic regions that encapsulate spatial and semantic dependencies. In contrast to prior methods that often discard high overlap predictions, we retain and exploit them as valuable supervisory signals. The set $\mathcal{R}$ comprises low-confidence queries that still carry weak relational cues and are utilized with reduced weighting.

\subsection{Overall Framework}
\label{subsec:Overall_framework}
We propose SIKD for incremental object detection with DETR-style detectors. At step $t$, the frozen model $\mathcal{M}^{t-1}$ runs on $\mathcal{D}^{t}$ to produce queries $\mathcal{Q}^{t-1}$ and predictions. Following Sec.~\ref{subsec:partition}, we partition the queries into stable anchors $\mathcal{A}$, symbiotic regions $\mathcal{S}$ and residual queries $\mathcal{R}$. The core of SIKD consist of two complementary distillation pathway. Spatial Symbiosis Distillation refines $\mathcal{S}$ with CFE while keeping $\mathcal{A}$ fixed and yields the enhanced set $\mathcal{Q}^{\mathrm{E}}$. SpSD then applies slot aligned, confidence weighted supervision by feeding $\mathcal{Q}^{\mathrm{E}}$ to the frozen old decoder and $\mathcal{Q}^{t-1}$ to the new decoder. This promotes instance-level consistency in space. Semantic Symbiosis Distillation aggregates the last layer decoder outputs of both models into confidence weighted, $L_2$ normalized prototypes for each old class.
It evaluates these prototypes in the old-class logit subspace and aligns their soft ranks to preserve semantic structure during adaptation. Together, SpSD and SeSD convert overlap-driven signals into reliable supervision and maintain a unified feature space across incremental tasks.

\subsection{Spatial Symbiosis Distillation}
\label{subsec:spsd}
Spatial Symbiosis Distillation (SpSD) leverages object symbiosis in the spatial dimension by exploiting co-occurrence and occlusion patterns that manifest as distinctive signatures in the feature space. As identified in Sec.~\ref{subsec:partition}, these symbiotic patterns concentrate in the overlap-driven set $\mathcal{S}$, while the set $\mathcal{A}$ provides semantically reliable anchors of old-class knowledge. Directly transferring knowledge from all old model queries causes two issues: It inflates old class confidence on unseen categories, and it suppresses responses to partially occluded old class objects, which leads to misclassification or background assignment. SpSD preserves anchor integrity and transforms symbiotic queries into anchor aligned representations, which maintains feature consistency across spatially related instances.

\noindent\textbf{Consistent Feature Enhancement.}
We aim to enhance spatially consistent features in symbiotic regions while preventing anchor drift. This requires modeling contextual relationships among queries to enable ambiguous symbiotic slots to aggregate evidence from reliable anchors. We refine queries using multi-head self-attention followed by an MLP while preserving anchors:
\begin{equation}
\begin{aligned}
\Delta\mathcal{Q}^{t-1}&=\mathrm{MLP}\big(\mathrm{MHA}(\mathcal{Q}^{t-1})\big),\\
\mathbf{q}_{i}^\mathrm{E}
&= \mathbf{q}^{t-1}_i
+ \mathbbm{1}_{\,i\notin\mathcal{A}\,}\;\Delta\mathbf{q}^{t-1}_i .
\end{aligned}
\label{eq:spsd_update}
\end{equation}
Here $\Delta\mathcal{Q}^{t-1}=[\Delta\mathbf{q}^{t-1}_1,\dots,\Delta\mathbf{q}^{t-1}_n]^\top$, $\Delta\mathbf{q}_{i}^{t-1}$ is the $i$-th row of $\Delta\mathcal{Q}^{t-1}$, and $\mathbbm{1}_{\,i\notin\mathcal{A}\,}=1$ if $i\notin\mathcal{A}$ and $0$ otherwise.

To guide this enhancement, we construct class-consistent prototypes from anchors. For each enhanced query $\mathbf{q}_{i}^{\mathrm{E}}$ with predicted class $y^{t-1}_i$, the confidence-weighted prototype is
\begin{equation}
\mathbf{p}_{i}
=\mathrm{norm}\!\left(
\frac{\sum_{j\in\mathcal{P}_{\mathcal{A}}(i)} s^{t-1}_j\,\mathbf{q}^{\mathrm{E}}_{j}}
     {\sum_{j\in\mathcal{P}_{\mathcal{A}}(i)} s^{t-1}_j+\varepsilon}
\right)\in\mathbb{R}^{d},
\label{eq:spsd_proto}
\end{equation}
where $\mathcal{P}_{\mathcal{A}}(i)=\{\,j\in\mathcal{A}\mid y^{t-1}_j=y^{t-1}_i,\ j\neq i\,\}$. We then align enhanced symbiotic queries to their prototypes with a cosine objective:
\begin{equation}
\mathcal{L}_{\mathrm{CFE}}
=\frac{1}{|\mathcal{S}|}
\sum_{i\in\mathcal{S}}
\Big(1-\mathrm{norm}(\mathbf{q}_{i}^{\mathrm{E}})^{\top}\mathbf{p}_{i}\Big).
\label{eq:ldqr}
\end{equation}
\newcommand{\VRule}{\vrule width 0.75pt}
\newcommand{\best}[1]{\textbf{#1}}

\begin{table*}[t]
  \caption{Experimental results (\%) on the COCO 2017 two-task settings. Best results are in \textbf{bold}. Methods marked with * use exemplars.}
  \label{tab:coco_two_stages}
  \centering
    \begin{tabular}{
      l !{\VRule} l !{\VRule} c !{\VRule} *{6}{c}
    }
    \toprule
    \makecell[c]{Setting} & \makecell[c]{Method} & \makecell[c]{Baseline}
    & $AP$ & $AP_{50}$ & $AP_{75}$ & $AP_S$ & $AP_M$ & $AP_L$ \\
    \midrule
    \multirow{12}{*}{\makecell[c]{70 + 10}}
    & LwF~\cite{li2017learning} & GFLv1  & 7.1 & 12.4 & 7.0 & 4.8 & 9.5 & 10.0 \\
    & RILOD~\cite{li2019rilod} & GFLv1  & 24.5 & 37.9 & 25.7 & 14.2 & 27.4 & 33.5 \\
    & SID~\cite{peng2021sid} & GFLv1  & 32.8 & 49.0 & 35.0 & 17.1 & 36.9 & 44.5 \\
    & ERD~\cite{feng2022overcoming} & GFLv1  & 34.9 & 51.9 & 37.4 & 18.7 & 38.8 & 45.5 \\
    & TLR~\cite{zhang2024learning} & GLIP  & 42.9 & 59.2 & 45.2 & 24.3 & 45.1 & 54.1 \\
    & CL-DETR*~\cite{liu2023continual} & Deformable DETR & 40.4 & 58.0 & 43.9 & 23.8 & 43.6 & 53.5 \\
    & DyQ-DETR*~\cite{zhang2025dynamic} & Deformable DETR & 42.4 & 60.4 & 46.3 & 24.5 & 45.7 & 57.5 \\
    & CL-DETR~\cite{liu2023continual} & Deformable DETR & 35.8 & 53.5 & 39.5 & 19.4& 41.5 & 46.1 \\
    & ACF~\cite{kang2023alleviating} & Deformable DETR & 37.6 & -- & -- & -- & -- & -- \\
    & DCA~\cite{zhang2025dca} & Deformable DETR & 41.3 & 59.2 & -- & -- & -- & -- \\
    & DyQ-DETR~\cite{zhang2025dynamic} & Deformable DETR & 39.5 & 56.4 & 43.1 & 22.5 & 43.1 & 53.0 \\
    & SIKD (Ours) & Deformable DETR & \best{44.3} & \best{62.9} & \best{47.8} & \best{28.2} & \best{47.7} & \best{59.5} \\
    \midrule
    \multirow{12}{*}{\makecell[c]{40 + 40}}
    & LwF~\cite{li2017learning} & GFLv1  & 17.2 & 25.4 & 18.6 & 7.9 & 18.4 & 24.3 \\
    & RILOD~\cite{li2019rilod} & GFLv1  & 29.9 & 45.0 & 32.0 & 15.8 & 33.0 & 40.5 \\
    & SID~\cite{peng2021sid} & GFLv1  & 34.0 & 51.4 & 36.3 & 18.4 & 38.4 & 44.9 \\
    & ERD~\cite{feng2022overcoming} & GFLv1  & 36.9 & 54.5 & 39.6 & 21.3 & 40.4 & 47.5 \\
    & TLR~\cite{zhang2024learning} & GLIP  & 40.4 & 57.4 & 43.9 & 23.3 & 44.7 & 54.5 \\
    & CL-DETR*~\cite{liu2023continual} & Deformable DETR & 42.0 & 60.1 & 45.9 & 24.0 & 45.3 & 55.6 \\
    & DyQ-DETR*~\cite{zhang2025dynamic} & Deformable DETR & 42.4 & 60.5 & 45.9 & 23.9 & 46.3 & 56.7 \\
    & CL-DETR~\cite{liu2023continual} & Deformable DETR & 39.2 & 56.1 & 42.6 & 21.0 & 42.8 & 52.6 \\
    & ACF~\cite{kang2023alleviating} & Deformable DETR & 39.8 & -- & -- & -- & -- & -- \\
    & DCA~\cite{zhang2025dca} & Deformable DETR & 42.8 & 58.4 & -- & -- & -- & -- \\
    & DyQ-DETR~\cite{zhang2025dynamic} & Deformable DETR & 41.4 & 59.7 & 44.9 & 24.1 & 45.2 & 54.3 \\
    & SIKD (Ours) & Deformable DETR & \best{43.3} & \best{61.7} & \best{46.7} & \best{26.4} & \best{46.6} & \best{57.0} \\
    \bottomrule
  \end{tabular}
\end{table*}

\noindent\textbf{Spatially Aligned Distillation.}
Let $\mathcal{Q}^{\mathrm{E}}=[\mathbf{q}^{\mathrm{E}}_1,\dots,\mathbf{q}^{\mathrm{E}}_n]^\top$ denote the enhanced query set. We feed $\mathcal{Q}^{\mathrm{E}}$ to the decoder of the frozen old model $\mathcal{M}^{t-1}$ and $\mathcal{Q}^{t-1}$ to the decoder of the new model $\mathcal{M}^{t}$, 
producing logits $\mathbf{z}^{\mathrm{E}}_{i},\hat{\mathbf{z}}^{t}_{i}\in\mathbb{R}^{m}$ and boxes $\mathbf{b}^{\mathrm{E}}_{i},\hat{\mathbf{b}}^{t}_{i}\in\mathbb{R}^{4}$, 
where $m=|\mathcal{C}^{1:t-1}|$ denotes the number of old classes. For anchors we distill semantics and geometry:
\begin{equation}
\begin{aligned}
\mathcal{L}_{A}
&=\frac{1}{|\mathcal{A}|}
\sum_{i\in\mathcal{A}}
\Big[\mathcal{L}_{\mathrm{KL}}\big(\hat{\mathbf{z}}^{t}_i,\mathbf{z}_{i}^{\mathrm{E}}\big)
+\lambda_{1}\mathcal{L}_{L1}\big(\hat{\mathbf{b}}^{t}_i,\mathbf{b}_{i}^{\mathrm{E}}\big)\\
&\phantom{=}
+\lambda_{2}\mathcal{L}_{\mathrm{GIoU}}\big(\hat{\mathbf{b}}^{t}_i,\mathbf{b}_{i}^{\mathrm{E}}\big)
\Big].
\end{aligned}
\label{eq:SpSD_A}
\end{equation}
Here $\mathcal{L}_{\mathrm{KL}}$ is the KL divergence on the old-class logit subspace, $\mathcal{L}_{L1}$ is the $\ell_1$ loss on box coordinates, and $\mathcal{L}_{\mathrm{GIoU}}$ is the generalized IoU loss.

To maintain consistency across decoder layers while handling prediction uncertainty, we employ layer-specific confidence weights at layer $\ell$ as:
\begin{equation}
    w_i^{(\ell)}=\frac{s^{t-1,(\ell)}_i}{\sum_{j} s^{t-1,(\ell)}_j+\varepsilon},
\end{equation}
where $s^{t-1,(\ell)}_i$ denotes the old model confidence for query $i$ at decoder layer $\ell$. The layer wise distillation objective enforces progressive feature alignment:
\begin{equation}
\mathcal{L}_{\mathrm{ID}}
=\sum_{\ell=1}^{L}
\sum_{i} w_i^{(\ell)}
\left\|\mathbf{z}_{i}^{\mathrm{E},(\ell)}-\hat{\mathbf{z}}^{t,(\ell)}_{i}\right\|_{2}^{2}.
\label{eq:spsd_layer}
\end{equation}

The overall spatial distillation objective combines both components:
\begin{equation}
\mathcal{L}_{\mathrm{SpSD}}
=\mathcal{L}_{\mathrm{A}}
+\alpha\mathcal{L}_{\mathrm{ID}},
\label{eq:spsd_total}
\end{equation}
where $\alpha$ balances the contributions from anchor distillation and layer-wise feature alignment. With this weighting, SpSD preserves spatial coherence by aligning symbiotic regions during incremental learning and reduces representation drift by turning object co-occurrence and occlusion into useful supervision.
\subsection{Semantic Symbiosis Distillation}
\label{subsec:sesd}
While SpSD preserves instance-level relationships, its effectiveness diminishes as query assignments shift during incremental training. Semantic Symbiosis Distillation (SeSD) addresses this limitation by transitioning to class-level structure preservation, maintaining the semantic topology of old classes through prototype-based alignment that remains robust to instance-level correspondence changes.

SeSD constructs stable class representations by aggregating features into confidence-weighted prototypes. For each old class $c\in\mathcal{C}^{1:t-1}$ across both model states:
\begin{equation}
\mathbf{p}^{\pi}_c = \mathrm{norm}\!\left(
\frac{\sum_{j\in\mathcal{P}^{\pi}(c)} s^{\pi}_j\mathbf{q}_{j}^{\pi,(L)}}
{\sum_{j\in\mathcal{P}^{\pi}(c)} s^{\pi}_j+\varepsilon}
\right),\quad \pi\in\{t-1,t\},
\label{eq:SeSD_proto}
\end{equation}
where $\pi$ indexes the model state ($t-1$ for old model, $t$ for new model), $\mathcal{P}^{\pi}(c)$ denotes the set of queries assigned to old class $c$ by model $\mathcal{M}^{\pi}$, and $s^{\pi}_j$ represents the corresponding confidence score derived from old-class logits.

To construct semantic relations among old classes, we project prototypes through the classifier heads $\mathcal{H}^{\pi}(\cdot)$ ($\pi \in \{t-1,t\}$) and normalize the resulting score vectors:

\begin{equation}
\tilde{\mathbf{s}}^{\pi}_c = \frac{\sigma\!\big(\mathcal{H}^{\pi}(\mathbf{p}^{\pi}_c)\big)_{1:m}}
{\max \left[\sigma\!\big(\mathcal{H}^{\pi}(\mathbf{p}^{\pi}_c)\big)_{1:m}\right]} \in \mathbb{R}^{m},
\label{eq:SeSD_scores}
\end{equation}
where $m = |\mathcal{C}^{1:t-1}|$ denotes the total number of old classes, and the new model is restricted to these old-class dimensions to prevent interference from new categories.

We align semantic structures by the distillation objective:
\begin{equation}
\mathcal{L}_{\mathrm{SeSD}}
=\frac{1}{(m)^{2}}\sum_{c=1}^{m}
\left\|\mathrm{rank}\!\big(\tilde{\mathbf{s}}^{t}_c\big)-\mathrm{rank}\!\big(\tilde{\mathbf{s}}^{t-1}_c\big)\right\|_{1},
\label{eq:SeSD_loss}
\end{equation}
where $\mathrm{rank}(\tilde{\mathbf{s}}^{t}_c)_k = \sum_{j=1}^{m}\sigma(-(\tilde{s}^t_{c,k}-\tilde{s}^t_{c,j}))$ computes the soft rank position of class $k$ within the score vector, representing its relative semantic ordering. This rank-based alignment preserves topological relationships independent of absolute confidence values~\cite{tao2020topology,liu2022model}, focusing on the essential semantic structure rather than magnitude variations.

This semantic alignment preserves the relative logit structure of old classes, anchoring $\mathcal{Q}^{t,(L)}$ to $\mathcal{Q}^{t-1,(L)}$ even when instance correspondences break down. By complementing spatial distillation with semantic structure preservation, SeSD stabilizes the feature space throughout tasks.

\subsection{Training Objective}
\label{subsec:training_obj}
The training objective combines our symbiotic distillation terms with the standard detection loss. This joint objective maintains performance across incremental steps:
\begin{equation}
\mathcal{L}_{\mathrm{total}}
=\underbrace{\mathcal{L}_{\mathrm{det}}+\mathcal{L}_{\mathrm{SpSD}}+\beta\mathcal{L}_{\mathrm{SeSD}}}_{\mathcal{L}_{\mathrm{model}}}
+\mathcal{L}_{\mathrm{CFE}}.
\label{eq:total_loss}
\end{equation}
Here $\mathcal{L}_{\mathrm{det}}$ is the standard DETR detection loss, $\mathcal{L}_{\mathrm{SpSD}}$ maintains spatial consistency through query-level alignment, $\mathcal{L}_{\mathrm{SeSD}}$ preserves semantic structure via prototype-based ranking, and $\mathcal{L}_{\mathrm{CFE}}$ enhances features in symbiotic regions. The model parameters are updated using $\mathcal{L}_{\mathrm{model}}$, while only the CFE module is optimized via $\mathcal{L}_{\mathrm{CFE}}$, with gradients isolated between these components. 
\section{Experiments}
\label{sec:Experiments}
\begin{table*}[t]
  \caption{Experimental results ($AP$ / $AP_{50}$, \%) on the COCO 2017 multi-task settings. Methods marked with * use exemplars.}
  \label{tab:mutiinc_results}
  \centering
  \setlength{\tabcolsep}{2pt}
  \begin{tabular}{l|c|cccc|cc}
    \toprule
    \multirow{2}{*}{Method} &
    \multirow{2}{*}{(1 -- 40)} &
    \multicolumn{4}{c|}{{40+10+10+10+10}} &
    \multicolumn{2}{c}{{40+20+20}} \\
    & &
    + (40 -- 50) &
    + (50 -- 60) &
    + (60 -- 70) &
    + (70 -- 80) &
    + (40 -- 60) &
    + (60 -- 80) \\
    \midrule
    RILOD~\cite{li2019rilod} & 45.7 / 66.3 & 25.4 / 38.9 & 11.2 / 17.3 & 10.5 / 15.6 & \phantom{0}8.4 / 12.5 & 27.8 / 42.8 & 15.8 / 4.0\phantom{0} \\
    SID~\cite{peng2021sid} & 45.7 / 66.3 & 34.6 / 52.1 & 24.1 / 38.0 & 14.6 / 23.0 & 12.6 / 23.3   & 34.0 / 51.8 & 23.8 / 36.5 \\
    ERD~\cite{feng2022overcoming} & 45.7 / 66.3 & 36.4 / 53.9 & 30.8 / 46.7 & 26.2 / 39.9 & 20.7 / 31.8 & 36.7 / 54.6 & 32.4 / 48.6 \\
    CL-DETR*~\cite{liu2023continual}  &  46.5 / 68.6   &   -- /  --  &   -- /  --  & -- / --& 28.1 / --\phantom{0.0}   &   -- /  --  & 35.3 / --\phantom{0.0}  \\
    ACF~\cite{kang2023alleviating}  &  48.0 / --\phantom{0.0}   &  39.1 / --\phantom{0.0}   &  35.4 / --\phantom{0.0}  & 32.0 / --\phantom{0.0}  & 30.3 / --\phantom{0.0}   &  39.3 / --\phantom{0.0}   &  36.6 / --\phantom{0.0}  \\
    DCA~\cite{zhang2025dca}  &  48.0 / 68.9   &  \textbf{44.0} / 61.2   &  41.1 / 56.5  & 39.2 / 53.8  & 37.2 / 49.6   &  42.7 / 59.6  & 40.3 / 54.1 \\
    SIKD (Ours) &   45.4 / 64.7  &   {43.6} / \textbf{62.3}    &  \textbf{41.1} / \textbf{59.9}  &  \textbf{39.8} / \textbf{57.8}   & \textbf{38.1} / \textbf{55.5}   &   \textbf{43.4} / \textbf{62.1}   &  \textbf{40.8} / \textbf{58.8} \\
    \bottomrule
    \end{tabular}
\end{table*}
\newcommand{\cmark}{\checkmark}

\begin{table*}[t]
  \caption{Ablations on COCO 2017 (70+10) with Deformable DETR. “All categories” reports the $AP$ of the final-phase model over all 80 categories. “Old categories” reports the $AP$ of the final-phase model on the 70 categories introduced in phase 1. “FPP” is the $AP$ difference between the phase-1 model and the final-phase model on those 70 categories (lower is better). Idx 5 corresponds to our method. }
  \label{tab:coco_ablation}
  \centering
   \begin{tabular}{l | ccc | ccc ccc ccc}
    \toprule
    \multirow{2}{*}{Idx} & \multirow{2}{*}{Raw KD} & \multirow{2}{*}{SpSD} & \multirow{2}{*}{SeSD} &
    \multicolumn{3}{c}{All categories $\uparrow$} &
    \multicolumn{3}{c}{Old categories $\uparrow$} &
    \multicolumn{3}{c}{FPP $\downarrow$} \\
     &  &  &  &
    $AP$ & $AP_{50}$& $AP_{75}$&
    $AP$ & $AP_{50}$& $AP_{75}$&
    $AP$ & $AP_{50}$&$AP_{75}$\\
    \midrule
    1 & &  & & 41.2 & 59.6 & 44.4 & 41.5 & 60.1 & 44.7 & 4.9 & 5.4 & 5.4\\
    2 &\cmark &  &  &  41.1 & 59.4 & 44.3 & 41.3 & 59.9 & 44.5 & 5.1 & 5.6 & 5.6\\
    3 & & \cmark &  & 42.3 & 60.8 & 45.8 & 42.8 & 61.6 & 46.2 & 3.6 & 3.9 & 3.9\\
    4 & &  & \cmark & 43.9 & 62.6 & 47.4 & 44.4 & 63.4 & 47.9 & 2.0 & 2.1 & 2.2\\
    5 & & \cmark & \cmark & 44.3 & 62.9 & 47.8 & 45.1 & 64.0 & 48.7 & 1.3 & 1.5 & 1.4\\
    \bottomrule
  \end{tabular}
\end{table*}

\subsection{Experimental Settings}
\label{subsec:Experimental_settings}
\noindent\textbf{Datasets and Evaluation Metrics.} Consistent with prior work~\cite{liu2023continual,kim2024sddgr,zhang2025dynamic,zhang2025dca}, we adopt the standard COCO 2017~\cite{lin2014microsoft} evaluation protocol and incremental-setting notation. We evaluate on COCO 2017, which contains 80 object categories with 118k training images and 5k validation images. Performance follows the standard COCO metrics. The primary metric, $AP$, is the average precision averaged over IoU thresholds from 0.50 to 0.95 in steps of 0.05. We also report ${AP}_{50}$ and ${AP}_{75}$ at single IoU thresholds of 0.50 and 0.75. Scale-specific scores ${AP}_{S}$, ${AP}_{M}$, and ${AP}_{L}$ evaluate small, medium, and large objects, where small means area $<32^{2}$ pixels, medium means $32^{2}\le\text{area}<96^{2}$, and large means area $\ge96^{2}$ pixels. For incremental settings denoted A + B, we follow prior work: the initial step contains A classes, and each subsequent step adds B new classes.

\noindent\textbf{Implementation Details.} We implement our method within MMDetection on Deformable DETR with a ResNet-50 backbone pre-trained on ImageNet. All experiments run on four RTX 4090 GPUs, and the basic training settings follow the official implementation~\cite{chen2019mmdetection}. We use fixed hyperparameters across all settings. Following prior work~\cite{wang2024kd}, we set $\lambda_{1}=5.0$ and $\lambda_{2}=2.0$. We set $\alpha=1.0$ and $\beta=6.0$ as our choices. To ensure comparability and reproducibility, we randomize the category order using the predefined random seed released with CL-DETR~\cite{liu2023continual} and adopt the resulting order. The pseudo-label selection threshold in each incremental phase is $0.4$, and the IoU threshold is $0.7$.
\subsection{Comparison with the State-of-the-Arts}
\label{subsec:Comparison_results}
\noindent\textbf{Two-task settings.} Table~\ref{tab:coco_two_stages} compares our method with prior state-of-the-art methods on the COCO 2017 two-task splits. 
Compared with DyQ-DETR and DCA, both implemented on Deformable DETR, our method improves $AP$ by {1.9} and {3.0} points on the 70+10 split and by {0.9} and {0.5} points on the 40+40 split. The corresponding $AP_{50}$ gains are {2.5} and {3.7} points on 70+10 and {1.2} and {3.3} points on 40+40. It also surpasses the GLIP-based TLR, with gains of {1.5} and {2.9} $AP$ points and {3.7} and {4.3} $AP_{50}$ points on 70+10 and 40+40, respectively.

\noindent\textbf{Multi-task settings.} Table~\ref{tab:mutiinc_results} reports results on COCO 2017 under the multi-task settings. In the initial base training phase, our implementation achieves slightly lower performance than some prior methods. However, our method establishes new state-of-the-art results in all subsequent incremental phases. Compared with DCA, under the 40+10+10+10+10 setting we improve $AP$ by 0.9 and $AP_{50}$ by 5.9 points. Under the 40+20+20 setting we improve $AP$ by 0.5 and $AP_{50}$ by 4.7. The improvements reflect superior retention of prior knowledge coupled with efficient acquisition of new concepts.
\subsection{Results and Analysis}
\label{subsec:Result_and_analyses}
\noindent\textbf{Results on DIOR dataset.} We further evaluate our method on the DIOR dataset~\cite{li2020object}. DIOR is a large-scale optical remote sensing benchmark with 20 categories and strong variation in scale, viewpoint, object density, and background, where co-occurrence and occlusion are common. We follow three two-task settings 10{+}10, 15{+}5, and 19{+}1 and report $AP_{50}$ as summarized in Table~\ref{tab:dior}. Compared with the replay-based method CL-DETR*, SIKD improves the All score by 6.2 points in the 10{+}10 setting, by 7.3 points in the 15{+}5 setting, and by 9.0 points in the 19{+}1 setting. 

\noindent\textbf{Ablation study of SIKD.} We evaluate the contribution of each component under the 70+10 setting, with results summarized in Table~\ref{tab:coco_ablation}. Idx 1 is the baseline, which adopts a standard pseudo-labeling strategy without any of our proposed modules. Idx 2 (Raw KD) incorporates high-IoU queries directly into the distillation process. Idx 3 employs only SpSD, and Idx 4 uses only SeSD. Idx 5 combines both SpSD and SeSD, representing our full SIKD method.

Compared to the baseline in Idx 1, directly integrating high-overlap queries in Index 2 causes a drop of 0.1 $AP$, suggesting that such queries introduce noise and bias when used naively. Using SpSD alone (Idx 3) improves $AP$ by 1.1 points, while SeSD alone (Idx 4) brings a more substantial gain of 2.7 points, highlighting the individual efficacy of each distillation pathway. The full SIKD model (Idx 5) achieves the best performance of $44.3$ $AP$, demonstrating that combining spatial and semantic distillation yields complementary benefits and the highest overall accuracy.
\begin{table}[t]
  \centering
  \caption{Experimental results (\%) on the DIOR dataset under the two-task settings. $AP_{50}$ is reported for Old, New, All and task-averaged (Avg). Best results are in \textbf{bold}. Methods marked with * use exemplars.}
  \label{tab:dior}
  \begin{tabular}{l|l|cccc}
    \toprule
    Setting & {Methods} & {Old} & {New} & \cellcolor{gray!18}{All} & {Avg} \\
    \midrule
    \multirow{3}{*}{10+10}
      & CL-DETR      & 42.2 & {63.9} & \cellcolor{gray!18}53.1 & 53.1 \\
      & CL-DETR*     & {64.4} & 61.5 & \cellcolor{gray!18}{63.0} & {63.0} \\
      & SIKD (Ours)    & \textbf{72.0} & \textbf{66.4} & \cellcolor{gray!18}\textbf{69.2} & \textbf{69.2} \\
    \midrule
    \multirow{3}{*}{15+5}
      & CL-DETR  & 43.0 & {63.8} & \cellcolor{gray!18}48.2 & 53.4 \\
      & CL-DETR* & {64.3} & 60.4 & \cellcolor{gray!18}{63.4} & {62.4} \\
      & SIKD (Ours) & \textbf{71.8} & \textbf{67.5} & \cellcolor{gray!18}\textbf{70.7} & \textbf{69.6} \\
    \midrule
    \multirow{3}{*}{19+1}
      & CL-DETR  & 47.6 & 44.0 & \cellcolor{gray!18}47.5 & 45.8 \\
      & {CL-DETR*} & {57.3} & {53.0} & \cellcolor{gray!18}{57.1} & {55.2} \\
      & SIKD (Ours) & \textbf{65.3} & \textbf{79.7} & \cellcolor{gray!18}\textbf{66.1} & \textbf{72.5} \\
    \bottomrule
  \end{tabular}
\end{table}

\begin{figure}[t]
  \centering
   \includegraphics[width=\linewidth]{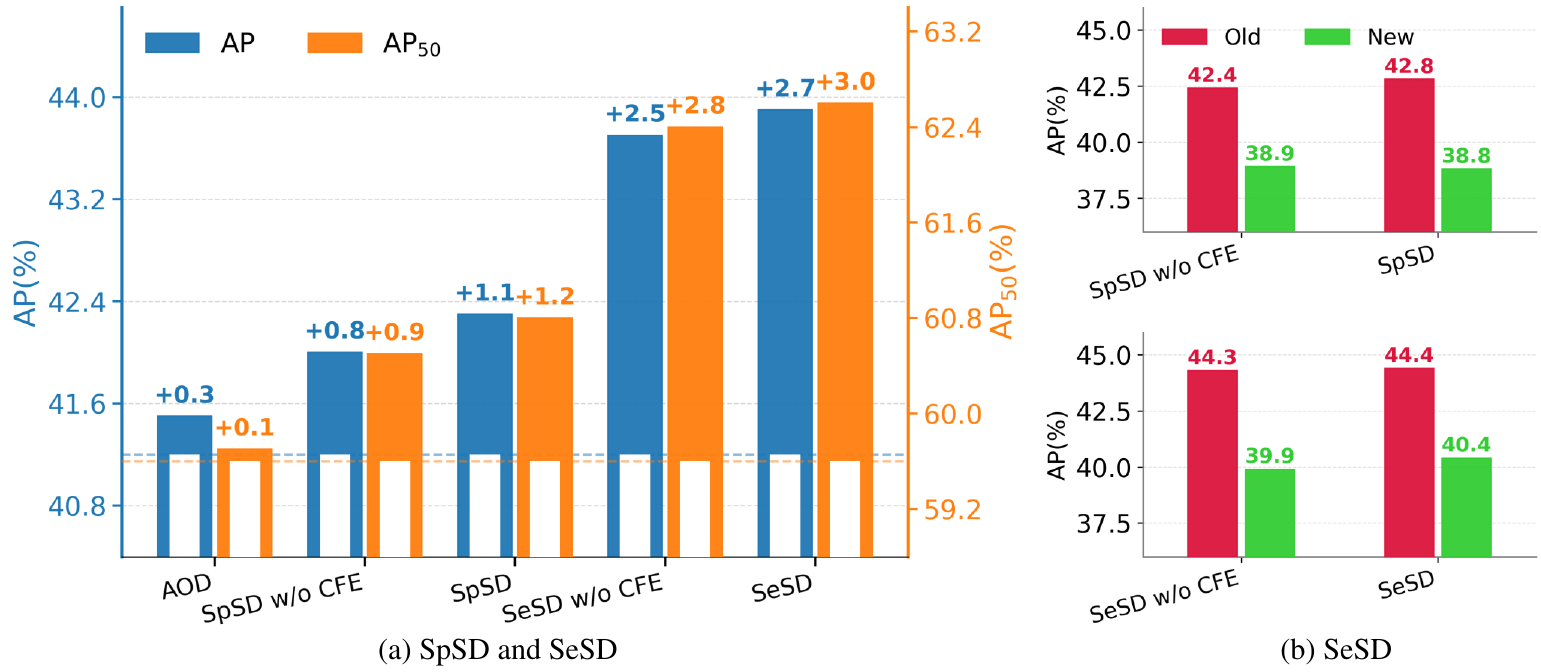}
   \caption{Ablations of SpSD and SeSD on COCO 2017 (70+10). In (a), the inner bars denote the baseline, with the full bars showing absolute values and labels indicating improvements over baseline.}
   \label{fig:ablation_grouped_bar_delta}
\end{figure}

\noindent\textbf{Analysis of SpSD and SeSD.} In Fig.~\ref{fig:ablation_grouped_bar_delta}(a), Anchor-Only Distillation (AOD) using Eq.~\ref{eq:SpSD_A} raises $AP$ by 0.3, indicating a modest gain. SpSD without CFE improves $AP$ by 0.8, showing that overlap-driven evidence is useful. Adding CFE to SpSD lifts the gain to 1.1, which further stabilizes instance-level consistency. SeSD without CFE improves $AP$ by 2.5, highlighting the value of preserving class-level topology. SeSD with CFE achieves 2.7, confirming that feature enhancement also benefits the class-level objective. In Fig.~\ref{fig:ablation_grouped_bar_delta}(b) and (c), CFE contributes 0.4 $AP$ to SpSD on the old task and 0.5 $AP$ to SeSD on the new task.

\begin{table}[t]
    \caption{Ablation of the balance weights on COCO 2017 (70+10). $AP$ is reported for Old, New, All and task-averaged (Avg).}
  \label{tab:coco_ablation_parameter}
  \centering
   \begin{tabular}{l | c c c  c}
    \toprule
    {Setting}  & {Old} & {New} & {All} & {Avg} \\
    \midrule
    $\beta=4$ & 44.7 & 39.1 & 44.0 &  41.9\\
    \rowcolor{gray!18}
    $\beta=6$ & 45.1 & 38.8 & 44.3 & {42.0}\\
    $\beta=8$ & {45.2} & 38.5 & {44.4} &  41.9\\
    \midrule
    $\alpha=0.1$ & 42.1 & 39.9 & 41.9 & 41.0\\
    \rowcolor{gray!18}
    $\alpha=1$ & 42.8 & 38.8 & 42.3 & {40.8}\\
    $\alpha=10$ & 43.2 & 34.9 & 42.2  & 39.1\\
    \bottomrule
    \end{tabular}
\end{table}

\noindent\textbf{Analysis of the balance weight.}
As shown in Table~\ref{tab:coco_ablation_parameter}, we ablate the balancing coefficients $\beta$ and $\alpha$ on COCO 2017 (70+10). For $\beta$, increasing the weight from 4 to 8 raises All $AP$ by 0.4 and Old $AP$ by 0.5, while reducing New $AP$ by 0.6. Setting $\beta=6$ yields the best task-averaged $AP$ (42.0) and provides a favorable trade-off between retention and plasticity, so we adopt $\beta=6$ in the main experiments. For $\alpha$, larger values bias training toward preserving old knowledge. Old $AP$ increases by 1.1 as $\alpha$ grows from 0.1 to 10, whereas New $AP$ drops by 5.0 and Avg $AP$ decreases by 1.9. Setting $\alpha=1$ achieves the highest All $AP$ (42.3) with a reasonable balance between old and new, and is therefore our default. 

\begin{figure}[t]
  \centering
   \includegraphics[width=\linewidth]{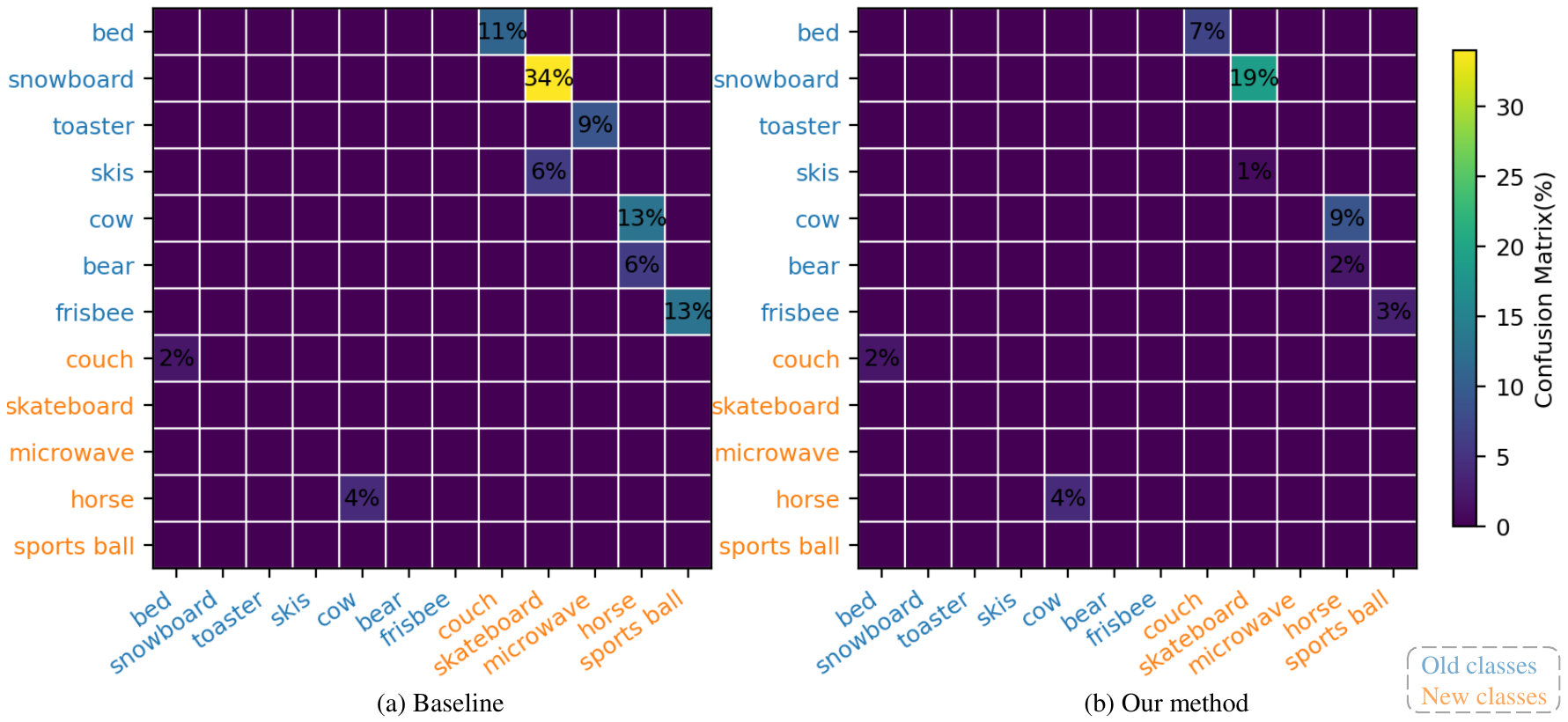}
   \caption{Comparison of confusion matrices for old and new Classes between the baseline and our method on COCO (70+10).}
   \label{fig:visual_confusion}
\end{figure}

\noindent\textbf{Visualizations.} As shown in Fig.~\ref{fig:visual_confusion}, we compare the confusion matrices between old and new categories for SIKD and the baseline on COCO 2017 (70+10). Relative to the baseline, SIKD reduces old-to-new confusion while not increasing new-to-old errors. These results support our design of maintaining a unified feature space across old and new classes, thereby mitigating the tendency of old-class knowledge to overfit toward new classes. The appendix includes additional experiments and visual analyses. Table~\ref{tab:efficiency} reports efficiency results, Table~\ref{tab:rank} provides further analysis of SeSD, Table~\ref{tab:dior_ablation_r_tao} ablates the hyperparameters $\gamma$ and $\tau$, and Table~\ref{tab:dior_muti} reports results under the multi-task setting. The appendix also contains Fig.~\ref{fig:visual_detection} and additional qualitative visualizations.

\section{Conclusion}
\label{sec:Conclusion}

We revisit IOD through object symbiosis and show that a unified feature space reduces confusion between old and new classes and curbs forgetting. SIKD models spatial symbiosis and semantic symbiosis. SpSD captures spatial symbiosis by refining symbiotic queries under anchor guidance and enforcing slot-aligned instance consistency. SeSD preserves semantic symbiosis by building confidence-weighted prototypes and aligning their ranks within the old-class subspace. Together, they convert overlap responses into reliable supervision and stabilize both spatial and semantic representations. Extensive experiments show consistent gains over state-of-the-art methods. In future work, we will explore using LLMs~\cite{cheng2026prompt,xu2026reasoning} to inject contextual priors to further strengthen symbiosis-aware distillation. 

\section*{Acknowledgments}
\label{sec:ackn}
This work was supported in part by the National Key R\&D Program of China under Grant No.2023YFA1008600, in part by the National Natural Science Foundation of China under
Grants 62576262, U22A2096, in part by the Key Research and Development Program of Shaanxi Province under grant 2024SF-YBXM-647, in part by the Fundamental Research Funds for the Central Universities under Grant QTZX25083, QTZX23042.

\section*{Impact Statement}
This paper presents work whose goal is to advance the field of Machine
Learning. There are many potential societal consequences of our work, none of
which we feel must be specifically highlighted here.

\bibliography{example_paper}

@article{li2017learning,
  title={Learning without forgetting},
  author={Li, Zhizhong and Hoiem, Derek},
  journal={IEEE transactions on pattern analysis and machine intelligence},
  volume={40},
  number={12},
  pages={2935--2947},
  year={2017},
  publisher={IEEE}
}

@inproceedings{li2019rilod,
  title={RILOD: Near real-time incremental learning for object detection at the edge},
  author={Li, Dawei and Tasci, Serafettin and Ghosh, Shalini and Zhu, Jingwen and Zhang, Junting and Heck, Larry},
  booktitle={Proceedings of the 4th ACM/IEEE Symposium on Edge Computing},
  pages={113--126},
  year={2019}
}

@article{peng2021sid,
  title={Sid: Incremental learning for anchor-free object detection via selective and inter-related distillation},
  author={Peng, Can and Zhao, Kun and Maksoud, Sam and Li, Meng and Lovell, Brian C},
  journal={Computer vision and image understanding},
  volume={210},
  pages={103229},
  year={2021},
  publisher={Elsevier}
}

@inproceedings{liu2023continual,
  title={Continual detection transformer for incremental object detection},
  author={Liu, Yaoyao and Schiele, Bernt and Vedaldi, Andrea and Rupprecht, Christian},
  booktitle={Proceedings of the IEEE/CVF Conference on Computer Vision and Pattern Recognition},
  pages={23799--23808},
  year={2023}
}

@inproceedings{feng2022overcoming,
  title={Overcoming catastrophic forgetting in incremental object detection via elastic response distillation},
  author={Feng, Tao and Wang, Mang and Yuan, Hangjie},
  booktitle={Proceedings of the IEEE/CVF conference on computer vision and pattern recognition},
  pages={9427--9436},
  year={2022}
}

@inproceedings{zhang2025dynamic,
  title={Dynamic Object Queries for Transformer-based Incremental Object Detection},
  author={Zhang, Jichuan and Li, Wei and Cheng, Shuang and Li, Yali and Wang, Shengjin},
  booktitle={ICASSP 2025-2025 IEEE International Conference on Acoustics, Speech and Signal Processing (ICASSP)},
  pages={1--5},
  year={2025},
  organization={IEEE}
}

@inproceedings{zhang2024learning,
  title={Learning task-aware language-image representation for class-incremental object detection},
  author={Zhang, Hongquan and Gao, Bin-Bin and Zeng, Yi and Tian, Xudong and Tan, Xin and Zhang, Zhizhong and Qu, Yanyun and Liu, Jun and Xie, Yuan},
  booktitle={Proceedings of the AAAI Conference on Artificial Intelligence},
  volume={38},
  pages={7096--7104},
  year={2024}
}

@inproceedings{wang2024kd,
  title={Kd-detr: Knowledge distillation for detection transformer with consistent distillation points sampling},
  author={Wang, Yu and Li, Xin and Weng, Shengzhao and Zhang, Gang and Yue, Haixiao and Feng, Haocheng and Han, Junyu and Ding, Errui},
  booktitle={Proceedings of the IEEE/CVF Conference on Computer Vision and Pattern Recognition},
  pages={16016--16025},
  year={2024}
}

@inproceedings{zhang2025dca,
  title={DCA: Dividing and Conquering Amnesia in Incremental Object Detection},
  author={Zhang, Aoting and Yang, Dongbao and Liu, Chang and Hong, Xiaopeng and Shang, Miao and Zhou, Yu},
  booktitle={Proceedings of the AAAI Conference on Artificial Intelligence},
  volume={39},
  number={9},
  pages={9851--9859},
  year={2025}
}

@inproceedings{kim2024sddgr,
  title={Sddgr: Stable diffusion-based deep generative replay for class incremental object detection},
  author={Kim, Junsu and Cho, Hoseong and Kim, Jihyeon and Tiruneh, Yihalem Yimolal and Baek, Seungryul},
  booktitle={Proceedings of the IEEE/CVF Conference on Computer Vision and Pattern Recognition},
  pages={28772--28781},
  year={2024}
}

@inproceedings{lin2014microsoft,
  title={Microsoft coco: Common objects in context},
  author={Lin, Tsung-Yi and Maire, Michael and Belongie, Serge and Hays, James and Perona, Pietro and Ramanan, Deva and Doll{\'a}r, Piotr and Zitnick, C Lawrence},
  booktitle={Computer vision--ECCV 2014: 13th European conference, zurich, Switzerland, September 6-12, 2014, proceedings, part v 13},
  pages={740--755},
  year={2014},
  organization={Springer}
}

@article{chen2019mmdetection,
  title={MMDetection: Open mmlab detection toolbox and benchmark},
  author={Chen, Kai and Wang, Jiaqi and Pang, Jiangmiao and Cao, Yuhang and Xiong, Yu and Li, Xiaoxiao and Sun, Shuyang and Feng, Wansen and Liu, Ziwei and Xu, Jiarui and others},
  journal={arXiv preprint arXiv:1906.07155},
  year={2019}
}

@article{kirkpatrick2017overcoming,
  title={Overcoming catastrophic forgetting in neural networks},
  author={Kirkpatrick, James and Pascanu, Razvan and Rabinowitz, Neil and Veness, Joel and Desjardins, Guillaume and Rusu, Andrei A and Milan, Kieran and Quan, John and Ramalho, Tiago and Grabska-Barwinska, Agnieszka and others},
  journal={Proceedings of the national academy of sciences},
  volume={114},
  number={13},
  pages={3521--3526},
  year={2017},
  publisher={National Academy of Sciences}
}

@article{wang2025ekpc,
  title={EKPC: Elastic Knowledge Preservation and Compensation for Class-Incremental Learning},
  author={Wang, Huaijie and Cheng, De and He, Lingfeng and Li, Yan and Li, Jie and Wang, Nannan and Gao, Xinbo},
  journal={arXiv preprint arXiv:2506.12351},
  year={2025}
}

@article{wang2025stpr,
  title={StPR: Spatiotemporal Preservation and Routing for Exemplar-Free Video Class-Incremental Learning},
  author={Wang, Huaijie and Cheng, De and Li, Guozhang and Xu, Zhipeng and He, Lingfeng and Li, Jie and Wang, Nannan and Gao, Xinbo},
  journal={arXiv preprint arXiv:2505.13997},
  year={2025}
}

@article{he2025harnessing,
  title={Harnessing Textual Semantic Priors for Knowledge Transfer and Refinement in CLIP-Driven Continual Learning},
  author={He, Lingfeng and Cheng, De and Wang, Huaijie and Wang, Nannan},
  journal={arXiv preprint arXiv:2508.01579},
  year={2025}
}

@article{he2025ckaa,
  title={CKAA: Cross-subspace Knowledge Alignment and Aggregation for Robust Continual Learning},
  author={He, Lingfeng and Cheng, De and Ma, Zhiheng and Wang, Huaijie and Zhang, Dingwen and Wang, Nannan and Gao, Xinbo},
  journal={arXiv preprint arXiv:2507.09471},
  year={2025}
}

@article{aljundi2019online,
  title={Online continual learning with maximal interfered retrieval},
  author={Aljundi, Rahaf and Belilovsky, Eugene and Tuytelaars, Tinne and Charlin, Laurent and Caccia, Massimo and Lin, Min and Page-Caccia, Lucas},
  journal={Advances in neural information processing systems},
  volume={32},
  year={2019}
}

@article{aljundi2019gradient,
  title={Gradient based sample selection for online continual learning},
  author={Aljundi, Rahaf and Lin, Min and Goujaud, Baptiste and Bengio, Yoshua},
  journal={Advances in neural information processing systems},
  volume={32},
  year={2019}
}

@article{zhou2024balanced,
  title={Balanced destruction-reconstruction dynamics for memory-replay class incremental learning},
  author={Zhou, Yuhang and Yao, Jiangchao and Hong, Feng and Zhang, Ya and Wang, Yanfeng},
  journal={IEEE Transactions on Image Processing},
  year={2024},
  publisher={IEEE}
}

@article{jung2020continual,
  title={Continual learning with node-importance based adaptive group sparse regularization},
  author={Jung, Sangwon and Ahn, Hongjoon and Cha, Sungmin and Moon, Taesup},
  journal={Advances in neural information processing systems},
  volume={33},
  pages={3647--3658},
  year={2020}
}

@inproceedings{yan2021dynamically,
  title={Der: Dynamically expandable representation for class incremental learning},
  author={Yan, Shipeng and Xie, Jiangwei and He, Xuming},
  booktitle={Proceedings of the IEEE/CVF conference on computer vision and pattern recognition},
  pages={3014--3023},
  year={2021}
}

@inproceedings{li2019learn,
  title={Learn to grow: A continual structure learning framework for overcoming catastrophic forgetting},
  author={Li, Xilai and Zhou, Yingbo and Wu, Tianfu and Socher, Richard and Xiong, Caiming},
  booktitle={International conference on machine learning},
  pages={3925--3934},
  year={2019},
  organization={PMLR}
}

@inproceedings{rebuffi2017icarl,
  title={icarl: Incremental classifier and representation learning},
  author={Rebuffi, Sylvestre-Alvise and Kolesnikov, Alexander and Sperl, Georg and Lampert, Christoph H},
  booktitle={Proceedings of the IEEE conference on Computer Vision and Pattern Recognition},
  pages={2001--2010},
  year={2017}
}

@inproceedings{wang2025gcd,
  title={GCD: Advancing Vision-Language Models for Incremental Object Detection via Global Alignment and Correspondence Distillation},
  author={Wang, Xu and Wang, Zilei and Lin, Zihan},
  booktitle={Proceedings of the AAAI Conference on Artificial Intelligence},
  volume={39},
  pages={8015--8023},
  year={2025}
}

@inproceedings{wangpsedet,
  title={PseDet: Revisiting the Power of Pseudo Label in Incremental Object Detection},
  author={Wang, Qiuchen and Chen, Zehui and Yang, Chenhongyi and Liu, Jiaming and Li, Zhenyu and Zhao, Feng},
  booktitle={The Thirteenth International Conference on Learning Representations},
  year={2025}
}

@inproceedings{mo2024bridge,
  title={Bridge past and future: Overcoming information asymmetry in incremental object detection},
  author={Mo, Qijie and Gao, Yipeng and Fu, Shenghao and Yan, Junkai and Wu, Ancong and Zheng, Wei-Shi},
  booktitle={European Conference on Computer Vision},
  pages={463--480},
  year={2024},
  organization={Springer}
}

@article{ren2016faster,
  title={Faster R-CNN: Towards real-time object detection with region proposal networks},
  author={Ren, Shaoqing and He, Kaiming and Girshick, Ross and Sun, Jian},
  journal={IEEE transactions on pattern analysis and machine intelligence},
  volume={39},
  number={6},
  pages={1137--1149},
  year={2016},
  publisher={IEEE}
}

@article{ge2021yolox,
  title={Yolox: Exceeding yolo series in 2021},
  author={Ge, Zheng and Liu, Songtao and Wang, Feng and Li, Zeming and Sun, Jian},
  journal={arXiv preprint arXiv:2107.08430},
  year={2021}
}

@inproceedings{girshick2015fast,
  title={Fast r-cnn},
  author={Girshick, Ross},
  booktitle={Proceedings of the IEEE international conference on computer vision},
  pages={1440--1448},
  year={2015}
}

@article{zhu2020deformable,
  title={Deformable detr: Deformable transformers for end-to-end object detection},
  author={Zhu, Xizhou and Su, Weijie and Lu, Lewei and Li, Bin and Wang, Xiaogang and Dai, Jifeng},
  journal={arXiv preprint arXiv:2010.04159},
  year={2020}
}

@inproceedings{liu2024grounding,
  title={Grounding dino: Marrying dino with grounded pre-training for open-set object detection},
  author={Liu, Shilong and Zeng, Zhaoyang and Ren, Tianhe and Li, Feng and Zhang, Hao and Yang, Jie and Jiang, Qing and Li, Chunyuan and Yang, Jianwei and Su, Hang and others},
  booktitle={European conference on computer vision},
  pages={38--55},
  year={2024},
  organization={Springer}
}

@inproceedings{kang2023alleviating,
  title={Alleviating catastrophic forgetting of incremental object detection via within-class and between-class knowledge distillation},
  author={Kang, Mengxue and Zhang, Jinpeng and Zhang, Jinming and Wang, Xiashuang and Chen, Yang and Ma, Zhe and Huang, Xuhui},
  booktitle={Proceedings of the IEEE/CVF International Conference on Computer Vision},
  pages={18894--18904},
  year={2023}
}

@article{zhou2024class,
  title={Class-incremental learning: A survey},
  author={Zhou, Da-Wei and Wang, Qi-Wei and Qi, Zhi-Hong and Ye, Han-Jia and Zhan, De-Chuan and Liu, Ziwei},
  journal={IEEE Transactions on Pattern Analysis and Machine Intelligence},
  year={2024},
  publisher={IEEE}
}

@inproceedings{dai2021up,
  title={Up-detr: Unsupervised pre-training for object detection with transformers},
  author={Dai, Zhigang and Cai, Bolun and Lin, Yugeng and Chen, Junying},
  booktitle={Proceedings of the IEEE/CVF conference on computer vision and pattern recognition},
  pages={1601--1610},
  year={2021}
}

@inproceedings{tian2019fcos,
  title={Fcos: Fully convolutional one-stage object detection},
  author={Tian, Zhi and Shen, Chunhua and Chen, Hao and He, Tong},
  booktitle={Proceedings of the IEEE/CVF international conference on computer vision},
  pages={9627--9636},
  year={2019}
}

@article{masana2022class,
  title={Class-incremental learning: survey and performance evaluation on image classification},
  author={Masana, Marc and Liu, Xialei and Twardowski, Bart{\l}omiej and Menta, Mikel and Bagdanov, Andrew D and Van De Weijer, Joost},
  journal={IEEE Transactions on Pattern Analysis and Machine Intelligence},
  volume={45},
  number={5},
  pages={5513--5533},
  year={2022},
  publisher={IEEE}
}

@inproceedings{li2024fcs,
  title={Fcs: Feature calibration and separation for non-exemplar class incremental learning},
  author={Li, Qiwei and Peng, Yuxin and Zhou, Jiahuan},
  booktitle={Proceedings of the IEEE/CVF conference on computer vision and pattern recognition},
  pages={28495--28504},
  year={2024}
}

@inproceedings{tao2020topology,
  title={Topology-preserving class-incremental learning},
  author={Tao, Xiaoyu and Chang, Xinyuan and Hong, Xiaopeng and Wei, Xing and Gong, Yihong},
  booktitle={European conference on computer vision},
  pages={254--270},
  year={2020},
  organization={Springer}
}

@article{liu2022model,
  title={Model behavior preserving for class-incremental learning},
  author={Liu, Yu and Hong, Xiaopeng and Tao, Xiaoyu and Dong, Songlin and Shi, Jingang and Gong, Yihong},
  journal={IEEE Transactions on Neural Networks and Learning Systems},
  volume={34},
  number={10},
  pages={7529--7540},
  year={2022},
  publisher={IEEE}
}

@article{li2020object,
  title={Object detection in optical remote sensing images: A survey and a new benchmark},
  author={Li, Ke and Wan, Gang and Cheng, Gong and Meng, Liqiu and Han, Junwei},
  journal={ISPRS journal of photogrammetry and remote sensing},
  volume={159},
  pages={296--307},
  year={2020},
  publisher={Elsevier}
}

@inproceedings{liu2023augmented,
  title={Augmented box replay: Overcoming foreground shift for incremental object detection},
  author={Liu, Yuyang and Cong, Yang and Goswami, Dipam and Liu, Xialei and Van De Weijer, Joost},
  booktitle={Proceedings of the IEEE/CVF international conference on computer vision},
  pages={11367--11377},
  year={2023}
}

@inproceedings{cheng2024disentangled,
  title={Disentangled prompt representation for domain generalization},
  author={Cheng, De and Xu, Zhipeng and Jiang, Xinyang and Wang, Nannan and Li, Dongsheng and Gao, Xinbo},
  booktitle={Proceedings of the IEEE/CVF Conference on Computer Vision and Pattern Recognition},
  pages={23595--23604},
  year={2024}
}

@inproceedings{xu2025adversarial,
  title={Adversarial domain prompt tuning and generation for single domain generalization},
  author={Xu, Zhipeng and Cheng, De and Jiang, Xinyang and Wang, Nannan and Li, Dongsheng and Gao, Xinbo},
  booktitle={Proceedings of the IEEE/CVF Conference on Computer Vision and Pattern Recognition},
  pages={18584--18595},
  year={2025}
}

@article{cheng2026prompt,
  title={Prompt Disentanglement via Language Guidance and Representation Alignment for Domain Generalization},
  author={Cheng, De and Xu, Zhipeng and Jiang, Xinyang and Li, Dongsheng and Wang, Nannan and Gao, Xinbo},
  journal={IEEE Transactions on Pattern Analysis and Machine Intelligence},
  year={2026},
  publisher={IEEE}
}

@inproceedings{xu2026reasoning,
  title     = {Reasoning-Driven Multimodal {LLM} for Domain Generalization},
  author    = {Xu, Zhipeng and Wang, Zilong and Jiang, Xinyang and Li, Dongsheng and Cheng, De and Wang, Nannan},
  booktitle = {The Fourteenth International Conference on Learning Representations},
  year      = {2026},
  url       = {https://openreview.net/forum?id=psJiUopUt7}
}

@inproceedings{
he2026task,
title={Task-Driven Subspace Decomposition for Knowledge Sharing and Isolation in LoRA-based Continual Learning},
author={Lingfeng He and De Cheng and Huaijie Wang and Xiaofeng Zhu and Xi Yang and Nannan Wang and Xinbo Gao},
booktitle={Forty-third International Conference on Machine Learning},
year={2026}
}
\bibliographystyle{icml2026}


\newpage
\appendix
\onecolumn
\section{Training pipeline for SIKD}
\label{sec:Training_Pipeline}
Algorithm~\ref{alg:sikd} illustrates the training details of our method.

\begin{algorithm}[H]
\caption{SIKD training in new task $t$}
\label{alg:sikd}
\begin{algorithmic}[1]

\REQUIRE Frozen old model $\mathcal{M}^{t-1}$, new model $\mathcal{M}^{t}$,
current dataset $\mathcal{D}^{t}$, thresholds $\gamma,\tau$, weights $\alpha,\beta$.

\FOR{each mini-batch $(X,\mathcal{Y}^t)$ in $\mathcal{D}^t$}

    \STATE \textbf{Old-model forward and statistics}
    \STATE $\mathcal{Q}^{t-1},\{\mathbf{z}_i^{t-1}\}_{i=1}^n$, and $\{\mathbf{b}_i^{t-1}\}_{i=1}^n \gets \mathcal{M}^{t-1}(X)$
    \FOR{$i=1$ to $n$}
        \STATE $s_i^{t-1} \gets \max \sigma(\mathbf{z}_i^{t-1})$, and $v_i^{t-1} \gets \max_j \mathrm{IoU}(\mathbf{b}_i^{t-1}, \mathbf{g}_j^{t})$
    \ENDFOR
    \STATE $\mathcal{A} \gets \{\,i \mid s_i^{t-1}\ge\gamma \land v_i^{t-1}<\tau\,\}$, $\mathcal{S} \gets \{\,i \mid v_i^{t-1}\ge\tau\,\}$, and $\mathcal{R} \gets \{\,i \mid s_i^{t-1}<\gamma \land v_i^{t-1}<\tau\,\}$

    \STATE \textbf{Consistent Feature Enhancement (CFE)}
    \STATE $\Delta\mathcal{Q}^{t-1} \gets \mathrm{MLP}(\mathrm{MHA}(\mathcal{Q}^{t-1}))$
    \FOR{$i=1$ to $n$}
        \IF{$i\in\mathcal{A}$}
            \STATE $\mathbf{q}^{\mathrm{E}}_i \gets \mathbf{q}^{t-1}_i$
        \ELSE
            \STATE $\mathbf{q}^{\mathrm{E}}_i \gets \mathbf{q}^{t-1}_i + \Delta\mathbf{q}^{t-1}_i$
        \ENDIF
    \ENDFOR
    \STATE $\mathcal{Q}^{\mathrm{E}} \gets [\mathbf{q}^{\mathrm{E}}_1,\dots,\mathbf{q}^{\mathrm{E}}_n]^\top$
    \STATE Build anchor-based prototypes $\{\mathbf{p}_i\}$ and compute $\mathcal{L}_{\mathrm{CFE}}$ as in Eq.~\eqref{eq:ldqr}

    \STATE \textbf{Spatial Symbiosis Distillation (SpSD)}
    \STATE $\{\mathbf{z}_i^{\mathrm{E}},\mathbf{b}_i^{\mathrm{E}}\}_{i=1}^n \gets \text{decoder of }\mathcal{M}^{t-1}(\mathcal{Q}^{\mathrm{E}})$, and $\{\hat{\mathbf{z}}_i^{t},\hat{\mathbf{b}}_i^{t}\}_{i=1}^n \gets \text{decoder of }\mathcal{M}^{t}(\mathcal{Q}^{t-1})$
    \STATE Compute anchor loss $\mathcal{L}_{A}$ using Eq.~\eqref{eq:SpSD_A}
    \STATE Compute layer-wise loss $\mathcal{L}_{\mathrm{ID}}$ with $w_i^{(\ell)}$ from Eq.~\eqref{eq:spsd_layer}
    \STATE $\mathcal{L}_{\mathrm{SpSD}} \gets \mathcal{L}_{A} + \alpha \mathcal{L}_{\mathrm{ID}}$

    \STATE \textbf{Semantic Symbiosis Distillation (SeSD)}
    \STATE Build confidence-weighted prototypes $\{\mathbf{p}_c^{t-1},\mathbf{p}_c^{t}\}$ for each old class $c$
    \STATE Obtain $\tilde{\mathbf{s}}_c^{t-1},\tilde{\mathbf{s}}_c^{t}$ and compute $\mathcal{L}_{\mathrm{SeSD}}$ via Eq.~\eqref{eq:SeSD_loss}

    \STATE \textbf{Total losses}
    \STATE $\hat{\mathcal{Y}}^t \gets \mathcal{M}^{t}(X)$
    \STATE $\mathcal{L}_{\mathrm{det}} \gets \mathcal{L}_{\mathrm{det}}(\hat{\mathcal{Y}}^t,\mathcal{Y}^t)$, and $\mathcal{L}_{\mathrm{model}} \gets \mathcal{L}_{\mathrm{det}} + \mathcal{L}_{\mathrm{SpSD}} + \beta \mathcal{L}_{\mathrm{SeSD}}$
    \STATE Update the new detector with $\mathcal{L}_{\mathrm{model}}$ and the CFE module with $\mathcal{L}_{\mathrm{CFE}}$, while blocking gradients from each loss to the other branch.
\ENDFOR

\STATE \textbf{Output:} updated new model $\mathcal{M}^{t}$

\end{algorithmic}
\end{algorithm}

\section{Effect of rank alignment in SeSD}
\label{sec:effect_rank}
Table~\ref{tab:rank} compares rank alignment with direct score distillation on COCO 2017 (70+10). Using rank alignment improves all metrics. Old rises from 42.4 to 44.4 (+2.0), New from 39.6 to 40.4 (+0.8), All from 42.1 to 43.9 (+1.8), and Avg from 41.0 to 42.4 (+1.4). These gains indicate that preserving the relative ordering of class scores is more robust than matching raw scores. Rank alignment reduces sensitivity to calibration and scale differences between models, which helps maintain the semantic topology of old classes while adapting to new ones.

\section{Sensitivity to \texorpdfstring{$\gamma$ and $\tau$}{gamma and tau}}
\label{sec:sensitivity_gamma_tau}
We analyze the sensitivity of SIKD to the two thresholds $\gamma$ and $\tau$ used in our symbiosis-aware query partitioning. As shown in Table~\ref{tab:dior_ablation_r_tao}, the performance is stable across a reasonable range of values. For $\gamma$, setting it to $0.4$ yields the best overall performance, while values $0.3$ and $0.5$ lead to only minor changes. For $\tau$, the default choice $\tau=0.7$ achieves the highest $AP_{50}$ (70.7), and both lower ($0.5$) and higher ($0.9$) thresholds result in small degradations. Overall, these results indicate that our method is not overly sensitive to threshold selection, and we use $\gamma=0.4$ and $\tau=0.7$ as default in all experiments.
\begin{table}[t]
  \centering
    \caption{SeSD ablation on COCO 2017 (70+10) comparing rank alignment with direct score distillation. $AP$ (\%) is reported for Old, New, All, and task averaged (Avg).}
  \label{tab:rank}
  \begin{tabular}{l|cccc}
    \toprule
    {Methods} & {Old} & {New} & \cellcolor{gray!18}{All} & {Avg} \\
    \midrule
      SeSD w/o rank      & 42.4 & {39.6} & \cellcolor{gray!18}42.1 & 41.0 \\
      SeSD     & {44.4} & 40.4 & \cellcolor{gray!18}{43.9} & {42.4} \\
    \bottomrule
  \end{tabular}
\end{table}

\begin{table}[t]
    \caption{Sensitivity analysis of thresholds $\gamma$ and $\tau$ on DIOR under the 15+5 incremental setting. $AP_{50}$ (\%) is reported for old classes (Old), new classes (New), and all classes (All).}
  \label{tab:dior_ablation_r_tao}
  \centering
   \begin{tabular}{l | c c c }
    \toprule
    {Setting}  & {Old} & {New} & {All} \\
    \midrule
    $\gamma=0.3$ & 70.9 & 66.9 & 69.9 \\
    \rowcolor{gray!18}
    $\gamma=0.4$ & 71.8 & 67.5 & 70.7 \\
    $\gamma=0.5$ & 71.4 & 68.2 &  70.6 \\
    \midrule
    $\tau=0.5$ & 71.2 & 67.9 & 70.4\\
    \rowcolor{gray!18}
    $\tau=0.7$ & 71.8 & 67.5 & 70.7\\
    $\tau=0.9$ & 71.1 & 68.2 &70.3\\
    \bottomrule
    \end{tabular}
\end{table}

\begin{table}[t]
  \centering
  \caption{Experimental results ($AP_{50}$, \%) on the DIOR dataset under the multi-task settings. Best results are in \textbf{bold}.}
  \label{tab:dior_muti}
  \begin{tabular}{lccc}
    \toprule
    {Methods} & {10+5+5} & {5+5+5+5} &  {10+2+2+2+2+2} \\
    \midrule
    CL-DETR      & 42.9 & {39.4} & 35.2 \\
    CL-DETR*     & {47.1} & 37.1 & {34.1} \\
    SIKD (Ours)  & \textbf{66.5} & \textbf{61.4} & \textbf{61.4} \\
    \bottomrule
  \end{tabular}
\end{table}
\begin{table}[t]
  \caption{Efficiency on COCO 2017 (70+10). $\mathrm{GFLOPs}^{1}$ denotes inference FLOPs; $\mathrm{GFLOPs}^{2}$ denotes training FLOPs per image.}
  \label{tab:efficiency}
  \centering
   \begin{tabular}{lccc}
    \toprule
    Method &
    $\mathrm{GFLOPs}^{1}$ &
    $\mathrm{GFLOPs}^{2}$ &
    \#Params (M) \\
    \midrule
    baseline & 125 & 500 & 82.41 \\
    Raw KD & 125 & 644 & 82.41 \\
    CFE & - & 2.169 & 1.05 \\
    SIKD & 125 & 694.169 & 83.46 \\
    \bottomrule
    \end{tabular}
\end{table}
\section{Multi-task evaluation on DIOR}
\label{sec:dior_multi}
We further evaluate SIKD under multi-task class-incremental settings on DIOR. Specifically, we consider three settings: 10{+}5{+}5, 5{+}5{+}5{+}5, and 10{+}2{+}2{+}2{+}2{+}2, and report the final-phase $AP_{50}$ in Table~\ref{tab:dior_muti}. Across all settings, SIKD consistently outperforms CL-DETR and CL-DETR*. Notably, the performance gap becomes larger as the number of incremental phases increases, suggesting that SIKD better mitigates forgetting and maintains effective knowledge transfer over longer learning sequences.

\begin{figure*}[t]
  \centering
   \includegraphics[width=\linewidth]{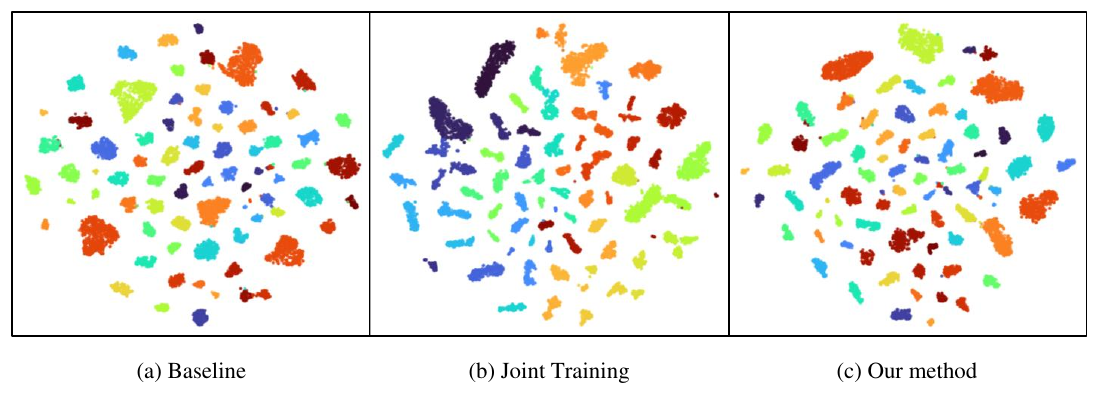}
    \caption{t-SNE visualization of object features on the COCO 2017 validation set. (a) Baseline after the final incremental task on COCO 2017 (70+10) setting, (b) Joint Training as an upper-bound model trained once on the union of old and new classes with full annotations (non-incremental), and (c) our method (SIKD) after the final incremental task on COCO 2017 (70+10) setting.
    }
   \label{fig:tsne}
\end{figure*}

\begin{figure}[t]
  \centering
   \includegraphics[width=0.65\linewidth]{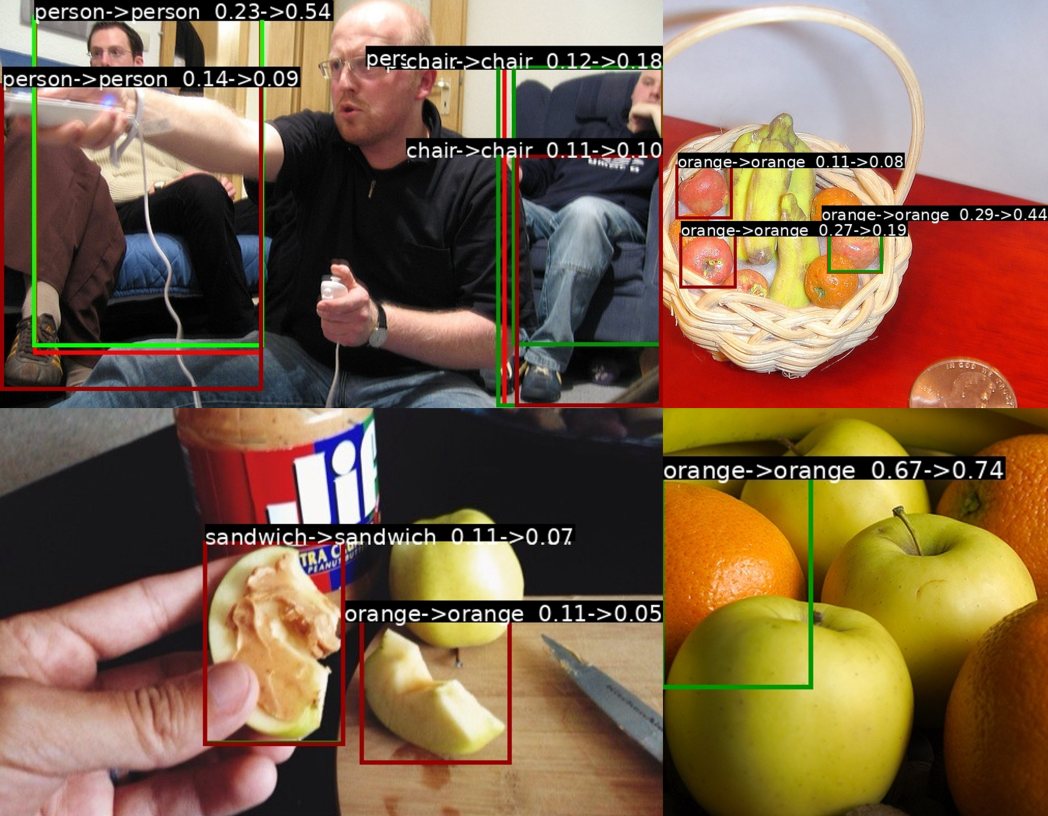}
   \caption{Visualization of the CFE module. For each detection, class id and confidence are shown as ``old→new'' with the left value before CFE and the right value after CFE.}
   \label{fig:visual_detection}
\end{figure}

\section{More visualization}
\label{sec:vis_more}
\subsection{t-SNE visualization}
\label{sec:tsne}
To qualitatively assess representation drift in class-incremental detection, we visualize the learned object features using t-SNE on the COCO 2017 validation set under the 70+10 setting in Figure~\ref{fig:tsne}. The baseline shows fragmented and less cohesive clusters, indicating unstable representations and aggravated old–new confusion after incremental updates. In contrast, our SIKD yields a more compact and well-structured embedding space that more closely resembles the joint-training reference. This suggests that the proposed symbiosis-inspired distillation better preserves old-class feature geometry while effectively incorporating new classes. Overall, this qualitative observation aligns with our quantitative results, supporting that SIKD alleviates catastrophic forgetting and improves feature consistency across incremental tasks.
\subsection{CFE module visualization}
 In Fig.~\ref{fig:visual_detection}, we visualize the per-detection transition in predicted class and confidence before and after applying CFE. Arrows indicate the mapping from the pre-CFE state to the post-CFE state.
 
\section{Efficiency of our method}
\label{sec:efficiency}
Table~\ref{tab:efficiency} reports inference and per-image training GFLOPs and parameter counts on COCO 2017 (70+10) with input size (1064$\times$800). All methods keep inference at 125 GFLOPs since no test-time modules are added. The baseline uses one forward of the frozen old model together with one forward–backward of the new detector, totaling 500 GFLOPs for training. Raw KD increases the training cost to 644 GFLOPs by adding an extra forward through the new decoder and the associated KD losses. SIKD retains these costs and further adds the training-only CFE block with both forward and backward, plus one additional forward of the frozen old decoder using enhanced queries for slot-aligned supervision, reaching 694.169 GFLOPs in training. CFE has no test-time cost, so inference remains unchanged.
\begin{figure*}[t]
  \centering
   \includegraphics[width=\linewidth]{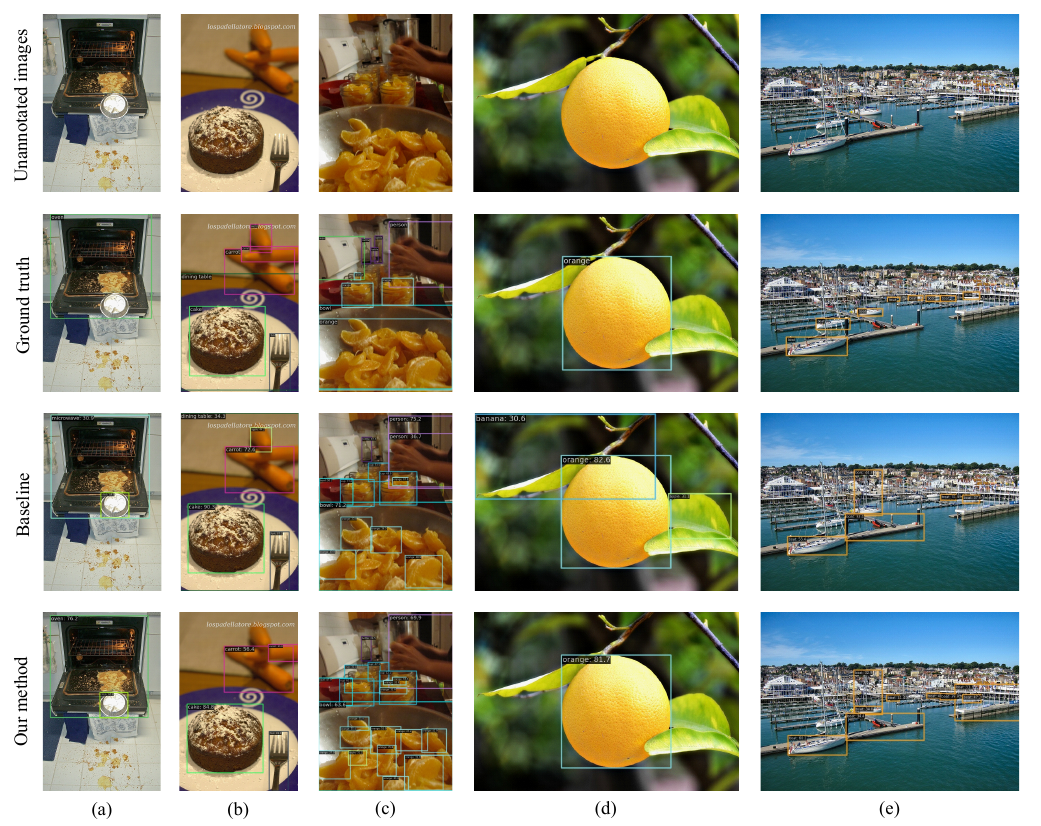}
    \caption{Qualitative visualization of detection predictions from the baseline and our method (SIKD) on COCO 2017 (70+10) setting. In (a), \textit{oven} is an old class and \textit{microwave} is a new class. In (b), \textit{carrot} is an old class and \textit{apple} is a new class. In (c), both \textit{orange} and \textit{person} are old classes. In (d), \textit{orange} is an old class, whereas \textit{banana} and \textit{apple} are new classes. In (e), \textit{boat} is an old class.}
   \label{fig:imgs_detection}
\end{figure*}
\section{Analysis of detection predictions}
\label{sec:img_visual}
As shown in Fig.~\ref{fig:imgs_detection}, we qualitatively compare detection predictions of the baseline and our SIKD on COCO 2017 under the 70+10 setting. In Fig.~\ref{fig:imgs_detection}(a), SIKD prevents the old class \textit{oven} from being confused with the visually similar new class \textit{microwave}. In Fig.~\ref{fig:imgs_detection}(b), SIKD avoids misclassifying the old class \textit{carrot} as the new class \textit{apple} and successfully detects hard instances that are missed by the baseline. In Fig.~\ref{fig:imgs_detection}(c), SIKD reduces forgetting of the old class \textit{orange}, whereas the baseline misses many orange objects. In Fig.~\ref{fig:imgs_detection}(d) and Fig.~\ref{fig:imgs_detection}(e), on images containing only old classes such as \textit{orange} or \textit{boat}, SIKD suppresses spurious predictions of new classes while reducing missed detections of old-class objects. These results demonstrate that SIKD simultaneously alleviates old–new confusion and preserves detection quality on old classes.


\end{document}